\documentclass{article}
\usepackage{ijcai26}

\usepackage{times}
\usepackage{soul}
\usepackage{url}

\usepackage[utf8]{inputenc}
\usepackage[small]{caption}
\usepackage{graphicx}
\usepackage{amsmath}
\usepackage{amsthm}
\usepackage{booktabs}

\usepackage{amsfonts}
\usepackage{xcolor}
\usepackage{xspace}

\usepackage{algpseudocode}

\usepackage{multirow}
\usepackage{colortbl}
\definecolor{lightgray}{gray}{0.95}

\newcommand{\method}[0]{\textsc{EventTSF}\xspace}

\usepackage{pifont}

\usepackage{makecell}

\usepackage{siunitx}
\usepackage{algorithm}

\usepackage[hidelinks]{hyperref}

\title{\textsc{EventTSF}: Event-Aware Non-Stationary Time Series Forecasting}

\author{
    Yunfeng Ge\textsuperscript{1,2},
    Ming Jin\textsuperscript{1}\thanks{Correspondence to Ming Jin (mingjinedu@gmail.com) and Shirui Pan (s.pan@griffith.edu.au).},
    Yiji Zhao\textsuperscript{3},
    Hongyan Li\textsuperscript{2},\\
    Bo Du\textsuperscript{1},
    Chang Xu\textsuperscript{4},
    Shirui Pan\textsuperscript{1}\footnotemark[1]
    \affiliations
    \textsuperscript{\rm 1}Griffith University, Australia
    \textsuperscript{\rm 2}Xidian University, China\\
    \textsuperscript{\rm 3}Yunnan University, China
    \textsuperscript{\rm 4}Microsoft Research Asia, China
}

\begin{document}

\maketitle

\begin{abstract}
Time series forecasting is vital in diverse sectors such as energy and transportation, where non-stationary dynamics are deeply intertwined with external events in other modalities such as texts.
However, incorporating natural language-based external events to improve non-stationary forecasting remains largely unexplored, as most approaches still rely on a single modality, resulting in limited contextual knowledge and model underperformance.
Enabling fine-grained multimodal interactions between temporal and textual data is challenged by two fundamental issues:
(1) the gap in modeling interactions among discrete external events and continuous time series in a unified framework; (2) classical uniform diffusion timestep ignores event-induced non-stationary variability, leading to imbalanced denoising difficulty across diffusion stages.
In this work, we propose event-aware non-stationary time series forecasting (\textbf{\method}), an autoregressive diffusion framework that integrates historical time series and textual events via step-wise diffusion.
To mitigate the imbalanced denoising difficulty of uniform timestep sampling, \method uses an event-aware flow-matching timestep conditioned on event semantics.
Extensive experiments on 7 synthetic and real-world datasets show that \method outperforms 12 non-stationary time series forecasting baselines, achieving average gains of $\mathbf{41.3\%}$ in probabilistic forecasting and $\mathbf{27.5\%}$ in deterministic forecasting across all evaluation metrics.

\end{abstract}


\section{Introduction}

\begin{figure}[t]
    \centering
    \includegraphics[width=1\linewidth]{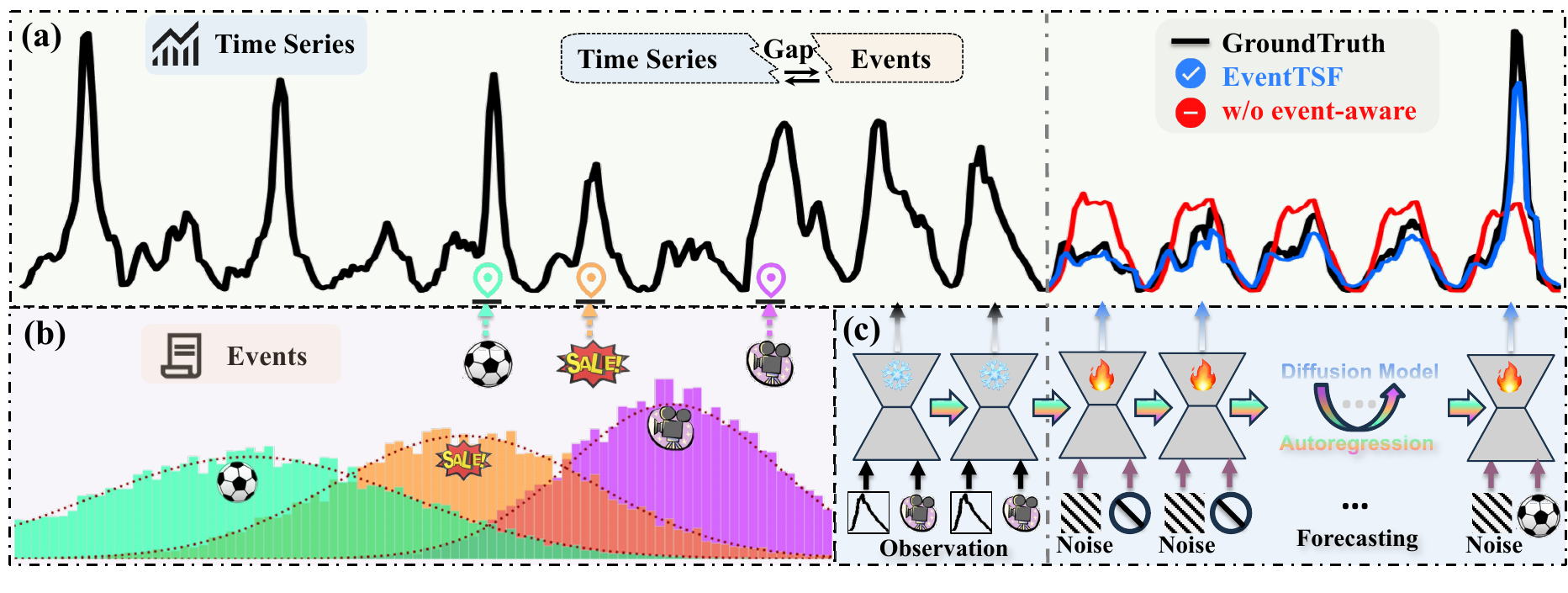}
    \caption{\textbf{Motivation.} Events induce distribution shifts.
    \textbf{(a)} Event-unaware forecasting degrades.
    \textbf{(b)} Events induce distinct temporal distributions.
    \textbf{(c)} \method jointly models events and time series.}
    \label{fig:motivation}
\end{figure}

Time series forecasting is critical in sectors such as energy, transportation, and meteorology~\cite{liang2024foundation}, where accurate forecasting enables effective decision making and resource management.
However, real-world forecasting faces longstanding challenges due to non-stationarity and distribution shifts~\cite{box1976time}, where the underlying time series distribution changes over time.
While recent advances, including normalization techniques (e.g., RevIN~\cite{kim2021reversible}), model decomposition (e.g., Koopa~\cite{liu2023koopa}), and online learning (e.g., OneNet~\cite{wen2023onenet}), have partially addressed these issues, they typically operate on a single modality, \textbf{overlooking that the change is entangled with external events from other modalities}, particularly texts.
This leaves valuable contextual knowledge largely unexploited, resulting in limited forecasting performance as evidenced in Figure~\ref{fig:motivation}~(a). The distinct segment-wise temporal distributions triggered by event type in Figure~\ref{fig:motivation}~(b) highlight the need for event-aware forecasting. To capture interactions between discrete textual events and continuous time series, we propose \method, an autoregressive generative architecture  (Figure~\ref{fig:motivation}~(c)).

In many real-world systems, upcoming events such as holidays or planned promotions are known in advance of the forecast horizon, providing actionable exogenous priors to anticipate distribution shifts that cannot be inferred from historical observations alone.
Recent efforts have explored multimodal integration for time series forecasting using large language models via exogenous information and agentic interactions~\cite{qiu2026dag,jiang2025explainable}.
However, these methods struggle to model non-stationarity driven by evolving temporal dynamics and complex event interactions.
\textbf{First}, existing methods utilize general and coarse-grained textual context for forecasting~\cite{zhang2025does}, but struggle to capture fine-grained event-induced distribution shifts.
\textbf{Second}, existing approaches rely on a deterministic event representation and produce point forecasts, limiting their ability to model event-induced uncertainty under non-stationary settings.
Recent advances in flow matching and autoregressive generation offer some promising tools for event-aware forecasting~\cite{lipman2022flow,liu2022flow}.

However, two key challenges remain unresolved for multimodal event-aware forecasting: \ding{182} \textbf{Fine-grained temporal-event interaction}. Existing methods~\cite{narasimhan2024time} rely on static global ``meta context'', which fails to synchronize and align fine-grained discrete events with continuous time series at the segment level. In contrast, modeling fine-grained temporal and event interactions helps capture cascading and lagged dependencies between time series and event dynamics.
Specifically, aligning chronologically ordered discrete event segments with the fixed-length time series windows enables step-wise event–series interaction, rather than conditioning on a single global context.
\ding{183} \textbf{Event-agnostic timestep sampling}.
Current diffusion-based time series forecasting models~\cite{ye2025non,yuan2024diffusion} typically sample diffusion timesteps from a fixed uniform distribution.
However, uniform sampling can skew denoising difficulty across diffusion stages, undermining stability.
Denoising difficulty is closely tied to the signal-to-noise ratio (SNR) of the denoiser’s learning target~\cite{choi2022perception}.
External events can induce non-stationary shifts in magnitude and variance, yielding heterogeneous SNR and thus varied event-dependent denoising difficulty.
Consequently, event-agnostic uniform timestep sampling cannot adapt to event-dependent dynamics, leading to imbalanced training difficulty and unstable denoiser optimization.

To address these challenges, we propose \method, a diffusion-based autoregressive architecture for event-aware non-stationary time series forecasting.
\method jointly models historical time series and textual events to generate probabilistic forecasts under non-stationary dynamics, maintaining a learnable multimodal hidden state at each step and incrementally incorporating event information.
To mitigate event-agnostic uniform diffusion timestep sampling, we introduce an event-aware mechanism that learns an event-conditioned reparameterization, mapping a base uniform sample to an event-dependent timestep via learnable distribution parameterized by event-conditioned networks.
This adaptive sampling strategy balances denoising difficulty across diffusion stages.
The implementation code is available\footnote{\url{https://github.com/WinfredGe/EventTSF}}. Our key contributions are summarized as follows:
\begin{itemize}
    \item \textbf{Paradigm Reformulation}. We formulate event-aware non-stationary time series forecasting as a paradigm that incorporates multimodal event information to better address distribution shifts in real-world forecasting.
    \item \textbf{Methodological Innovation}. We develop \method, an autoregressive diffusion model that enables fine-grained event–series alignment and mitigates denoising imbalance via event-aware timestep sampling.
    \item \textbf{Superior Performance}. \method consistently improves forecasting performance across diverse scenarios, improving probabilistic forecasting by $41.3\%$ and deterministic forecasting by $27.5\%$ on average overall.
\end{itemize}


\section{Definition}
\label{definition}

Given a univariate time series $\mathbf{X}_{1:L} = \{x_l\}_{l=1}^L$ of length $L$, we define a set of $N$ chronologically ordered fine-grained textual event embeddings $\mathcal{C} = \{ \mathbf{c}_{1}, \mathbf{c}_2, \ldots, \mathbf{c}_{s}, \ldots, \mathbf{c}_N\}$. Each event embedding $\mathbf{c}_{s}$ corresponds to an aligned non-overlapping consistent time segment $\mathbf{x}_{s}=\{x_l\}_{l=i_{s}}^{j_{s}} \subset \mathbf{X}_{1:L}$, where $i_{s}$ and $j_{s}$ denote the start and end timestamps of event segment $s$. The multimodal time series datasets is defined as $\mathcal{D} = \{(\mathbf{x}_s, \mathbf{c}_s)\}_{s=1}^N$, satisfying the constraints: (1) $\bigcup_{s=1}^N \mathbf{x}_s \subseteq \mathbf{X}_{1:L}$(complete coverage), and (2) $\mathbf{x}_s \cap \mathbf{x}_{s'} = \emptyset$ for all $s \neq s'$(non-overlapping segments).

We address event-aware non-stationary time series forecasting, where the underlying time series distribution exhibits event-conditioned shifts across different event categories,
$$P(\mathcal{X} \mid \mathbf{c}_i) \neq P(\mathcal{X} \mid \mathbf{c}_j), \quad \text{if} \quad \mathbf{c}_i \neq \mathbf{c}_j.$$
Given $p$ historical time series and event pairs $\mathcal{H}_\mathbf{x} = \{\mathbf{x}_l\}_{l=s-p+1}^{s}$ and $\mathcal{H}_\mathbf{c} = \{\mathbf{c}_l\}_{l=s-p+1}^{s}$, and $q$ future event descriptions $\mathcal{F}_\mathbf{c} = \{\mathbf{c}_l\}_{l=s+1}^{s+q}$, the objective is to forecast the following $q$ future time series predictions $\hat{\mathcal{F}}_\mathbf{x} = \mathcal{G}_\theta(\mathcal{H}_\mathbf{x}, \mathcal{H}_\mathbf{c}, \mathcal{F}_\mathbf{c})$, where $\hat{\mathcal{F}}_\mathbf{x}$ is the estimated $\mathcal{F}_\mathbf{x} = \{\mathbf{x}_l\}_{s+1}^{s+q}$; $\mathcal{G}_\theta$ denotes an event-aware forecasting model that captures non-stationary dynamics conditioned on external events.


\section{Methodology}
\begin{figure*}
    \centering
    \includegraphics[width=\linewidth,height=0.4\textheight,keepaspectratio]{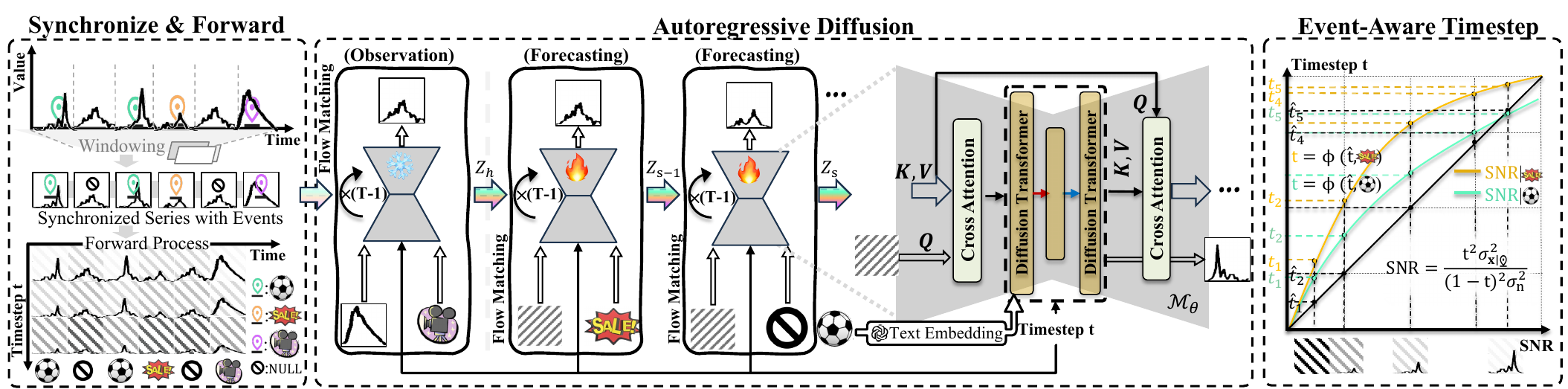}
    \caption{Overview of \method. \textbf{(Left)} Continuous time series is synchronized with discrete events via fixed-length windowing. The forward diffusion process injects noise with event-aware timestep. \textbf{(Middle)} An autoregressive diffusion architecture jointly models events and time series. At each step, it conditions on observed time series history and aligned event embedding, and progressively denoising future segments in an autoregressive manner. \textbf{(Right)} Since SNR characterizes denoising difficulty and varies across events, we replace uniform timestep sampling by reparameterizing the uniformly sampled base timestep $\hat{t}$ into an event-aware timestep $t=\phi(\hat{t},\mathbf{c})$.}
    \label{fig:architecture}
\end{figure*}

As shown in Figure~\ref{fig:architecture}, we introduce \method, an autoregressive diffusion architecture designed for event-aware non-stationary time series forecasting.
\method addresses events and time series synchronization and joint modeling.
\method integrates event-aware diffusion timestep, allowing event information to modulate the denoising process.

\subsection{Multimodal Autoregressive Diffusion}

\noindent\textbf{Event Synchronization}.
Discrete events often exert influence on continuous time series in real-world forecasting. For example, traffic demand often surges during public celebrations; network bandwidth usage can spike during highly anticipated live broadcasts; and product sales rise during planned promotional campaigns.
However, due to the lack of an effective joint representation, continuous time series and stochastic discrete events have been studied as two separate research topics: time series and temporal point process.
To bridge this modality gap, we propose event synchronization, which constructs a unified event-series representation under a shared temporal granularity.
Specifically, as illustrated in the left part of Figure~\ref{fig:architecture}, we partition the time series into non-overlapping segment $\mathbf{x}_{s}$. according to selected window size, and align each segment with an event embedding $\mathbf{c}_{s}$ that occurs within the same interval.
If no event occurs in a segment window, we assign a $\textsc{null}$ event.
Synchronized event–time series pairs $\mathcal{D} = \{(\mathbf{x}_s, \mathbf{c}_s)\}_{s=1}^N$ are then used as inputs to our autoregressive diffusion model.
Specifically, this model is trained on the synchronized multimodal segments produced by the forward flow matching process in Section~\ref{sectionflowmatching}.

\noindent\textbf{Autoregressive Diffusion}.
The integration of autoregressive modeling and diffusion processes has proven effective for temporal data modeling~\cite{chen2024diffusion}.
We design an autoregressive diffusion model \method for event-aware time series forecasting tasks, as illustrated in the middle of Figure~\ref{fig:architecture}.
This model regards the probabilistic time series forecasting as an autoregressive fine-grained segments generation.
Specifically, the model generates the segements $\mathbf{x}_s$ at forecasting step $s$ via a conditional diffusion denoising process.
This denoising is initialized by a latent state $\mathbf{Z}_{h}$ that summarizes historical time series and event observations.
The hidden state $\mathbf{Z}_{s}$ is updated at each step in an autoregressive manner through interactions with temporal segments and event semantics across the forecasting horizon.

During the training stage, we autoregressively optimize a flow-matching denoiser $\mathcal{M}_\theta$ conditioned on latent state $\mathbf{Z}_{s-1}$, noised time series segments $\mathbf{x}_{s}^{t}$, event-aware diffusion timestep $t$, and textual event embedding $\mathbf{c}_s$ encoded by a frozen OpenAI \textsl{text-embedding-3} model.
Specifically, at each autoregressive step $s$ and diffusion timestep $t$, the flow matching denoiser learns the velocity field $\mathbf{v}_s^t$ and updated latent state $\mathbf{Z}_{s}$ from
$\mathbf{v}_s^t, \mathbf{Z}_{s}=\mathcal{M}_\theta(\mathbf{Z}_{s-1},\mathbf{x}_s^t,\mathbf{c}_s,t)$ conditioned on the latent state $\mathbf{Z}_{s-1}$ and the event representation $\mathbf{c}_s$.
The denoiser $\mathcal{M}_\theta$ is shared across all autoregressive steps, while step specific dynamics are captured through the evolving latent state $\mathbf{Z}_{s}$ and event representation $\mathbf{c}_s$.
This design enables fine-grained relationship modeling between synchronized temporal patterns and event semantics.
The detailed training procedure is summarized in Algorithm~\ref{alg:autoregressive_training_concise}.

During the inference stage, we first compute the historical state $\mathbf{Z}_h$ from the clean historical time series $\mathcal{H}_x$ and the corresponding historical events $\mathcal{H}_c$, which is denoted in Section~\ref{definition}.
During forecasting, the historical state $\mathbf{Z}_h$ serves as the initial state $\mathbf{Z}_{0}$. At each autoregressive step $s$, the model takes a noise input $\mathbf{n}_{s}$ and an exogenous event $\mathbf{c}_{s}$ to generate the denoised segment $\mathbf{x}_{s}$ via the learned flow matching denoiser $\mathcal{M}_\theta(\mathbf{Z}_{s-1}, \mathbf{n}_{s}, \mathbf{c}_{s}, t)$, coupled with an \text{ODE} solver.
The complete architecture of the denoiser is provided in Appendix~\ref{app:mudit}, and the sampling procedure is detailed in Algorithm~\ref{alg:validation_sampling}.
Overall, \method autoregressively incorporates dynamic event descriptions, enabling flexible and event-aware forecasting in evolving environments.

\subsection{Flow Matching with Event-Aware Timestep}
\label{sectionflowmatching}
\noindent\textbf{Flow Matching}. We use the flow matching~\cite{liu2022flow,ge2025t2s} to jointly train textual events and time series. As illustrated in the left part of Figure~\ref{fig:architecture}, the forward diffusion process injects noise according to an event-aware diffusion timestep $t$. At each autoregressive step $s$ and diffusion timestep $t \in [0, 1]$,
flow matching constructs data trajectories that transform samples from a simple prior distribution $p(\mathbf{x}_{s}^{0}) \sim \mathcal{N}(\mathbf{0}, \mathbf{I})$ into the target data distribution $p(\mathbf{x}_{s}^{1})$.
An optimal transport path, realized through Rectified Flow~\cite{liu2022flow}, guides this transformation process:
\begin{equation}
    \label{addnoise}
    \mathbf{x}_{s}^{t} = (1-t) \mathbf{x}_{s}^{0} + t \mathbf{x}_{s}^{1}.
\end{equation}
The instantaneous velocity vector field along this path $\mathbf{v}_{s}^t$ with respect to diffusion timestep $t$ is given by:
\begin{equation}
\label{eq:velocity}
    \mathbf{v}_{s}^t = \frac{d\mathbf{x}_{s}^t}{dt} = \mathbf{x}_{s}^{1} - \mathbf{x}_{s}^{0}.
\end{equation}

A neural network $\mathbf{v}_{\theta}(\mathbf{x}_{s}^{t}, \mathbf{Z}_{s-1}, \mathbf{c}_{s}, t)$ approximates the target vector field $\mathbf{v}_{s}^t$. To enhance generation quality, $\mathbf{v}_{\theta}$ is designed to integrate the conditional information~\cite{lipman2022flow}. In our settings, this conditional information includes the historical state $\mathbf{Z}_{s-1}$ of the time series and the textual condition $\mathbf{c}_{s}$.
During sampling, the model generates $\hat{\mathbf{x}}_{s}^{1}$ by first sampling noise $\mathbf{x}_{s}^{0} \sim \mathcal{N}(\mathbf{0}, \mathbf{I})$ and then solving the learned velocity field $\mathbf{v}_{\theta}$ via a numerical \text{ODE} solver.

\noindent\textbf{Event-Aware Timestep}.
Real-world time series exhibit event-dependent dynamics, and different events can induce distinct signal magnitudes and variances, leading to heterogeneous denoising difficulty.
Denoising difficulty is closely tied to the $\mathrm{SNR}$ of the denoiser’s learning target~\cite{choi2022perception}; accordingly, we quantify denoising difficulty via $\mathrm{SNR}$.
Under the standard noise injection in Equation~\ref{addnoise}, the $\mathrm{SNR}$ at diffusion timestep $t$ is given by:
\begin{equation}
\mathrm{SNR}(t;\mathbf{c})
=\frac{t^{2}\sigma_{\mathbf{x}\mid \mathbf{c}}^{2}}{(1-t)^{2}\sigma_{\mathbf{n}}^{2}} ,
\end{equation}

\noindent where $\sigma_{\mathbf{x}\mid \mathbf{c}}^{2}$ and $\sigma_{\mathbf{n}}^{2}$ denote the variances of the clean signal and noise, respectively.
Notably, $\sigma_{\mathbf{x}\mid \mathbf{c}}^{2}$ can vary across event conditions, implying that the effective denoising difficulty at the same $t$ may differ under different events.

However, the diffusion timestep $t$ is typically sampled from a condition-agnostic distribution, such as uniform distribution $t \sim \mathcal{U}(0,1)$.
Such uniform sampling fails to allocate denoising difficulty appropriately; it ignores event-induced heterogeneity, resulting in uneven difficulty allocation for the denoiser learning.
As illustrated in the right panel of Figure~\ref{fig:architecture}, $\mathrm{SNR}$ varies nonlinearly with the diffusion timestep $t$ increase, indicating that denoising difficulty varies nonlinearly across different diffusion stages. Moreover, the timestep–$\mathrm{SNR}$ curve is event-dependent, resulting in heterogeneous denoising difficulty under different event conditions.

\begin{algorithm}[t]
\caption{Training}
\label{alg:autoregressive_training_concise}
\begin{algorithmic}[1]
\Require Training samples $(\mathbf{x}_s, \mathbf{c}_s)$; number of time series and event pairs per sample $S$; learnable initial state $\mathbf{Z}_0$; flow matching model $\mathcal{M}_\theta$; the learnable event-aware reparameterization function $\phi$; learning rate $\eta$; optimal transport path $OT$; Uniform distribution $\mathcal{U}(0, 1)$.
\State Initialize predicted velocity set $\hat{\mathcal{V}} \gets \{\}$.
\State Initialize groundtruth velocity set $\mathcal{V} \gets \{\}$.
\For{$s = 1, \dots, S$}
    \State Sample $\hat{t} \sim \mathcal{U}(0, 1)$
    \State $t \gets \phi(\hat{t}, \mathbf{c}_s)$
    \State Define $\mathbf{x}_{s}^{{t}}=OT(\mathbf{x_s},t)$ and $\mathbf{v}_{s}^{t}=\frac{d\mathbf{x}_{s}^{t}}{d\mathbf{t}}$
    \State Union Update $(\mathbf{Z}_s, \mathbf{\hat{v}}_{s}^{t}) = \mathcal{M}_\theta(\mathbf{Z}_{s-1},\mathbf{x}_{s}^{t},\mathbf{c}_{s},t)$
    \State $\mathcal{V} \gets \mathcal{V} \cup \{\mathbf{v}_{s}^{t} \}$
    \State $\hat{\mathcal{V}} \gets \hat{\mathcal{V}} \cup \{\mathbf{\hat{v}}_{s}^{t}\}$
\EndFor
\State $\mathcal{L} = \text{MSE}(\mathcal{V},\hat{\mathcal{V}})$
\State $\theta \gets \theta - \eta \nabla_\theta \mathcal{L}$

\end{algorithmic}
\end{algorithm}

To address these issues, we reparameterize uniformly sampled base timestep via an event-conditioned mapping, such that the effective denoising difficulty is more evenly allocated in an event-modulated manner.
Specifically, we map a base timestep $\hat{t} \in [0,1]$ drawn from a Uniform distribution $\hat{t} \sim \mathcal{U}(0,1)$ into an event-adaptive timestep $t \in [0,1]$, whose distribution follows a Beta law $t | \mathbf{c} \sim \mathrm{Beta}(\alpha(\mathbf{c}),\beta(\mathbf{c}))$:
\begin{equation}
t = \phi(\hat{t}, \mathbf{c}) = F^{-1}_{\mathrm{Beta}(\alpha(\mathbf{c}),\beta(\mathbf{c}))}(\hat{t}),
\end{equation}
\noindent where $\phi$ is the event-aware reparameterization function parameterized by $\Theta$, and $F^{-1}_{\alpha,\beta}(\cdot)$ is the inverse cumulative distribution function of a Beta distribution with parameters $\alpha$ and $\beta$.
This design ensures: (i) $t \in [0,1]$, (ii) $\phi$ is strictly monotone increasing in $\hat{t}$, preserving the intrinsic ordering of diffusion steps, and (iii) the mapping shape can be flexibly adjusted through $(\alpha,\beta)$ to realize condition-dependent adaptive schedules.

The Beta distribution parameters are predicted from the event representation $\mathbf{c}$ via two lightweight linear networks, $f_{\Theta_{\alpha}}$ and $g_{\Theta_{\beta}}$, followed by $\mathrm{softplus}(\cdot)$ activation:
\begin{equation}
\begin{aligned}
\alpha(\mathbf{c}) &= \mathrm{softplus}\big(f_{\Theta_{\alpha}}(\mathbf{c})\big) + \epsilon, \\
\beta(\mathbf{c})  &= \mathrm{softplus}\big(g_{\Theta_{\beta}}(\mathbf{c})\big) + \epsilon,
\end{aligned}
\end{equation}
\noindent where $f_{\Theta_{\alpha}}(\cdot)$ and $g_{\Theta_{\beta}}(\cdot)$ denote two separate lightweight learnable linear networks parameterized by $\Theta_{\alpha}$ and $\Theta_{\beta}$, respectively;
$\mathrm{softplus}(\cdot)$ enforces $\alpha(\mathbf{c}),\beta(\mathbf{c})>0$; and $\epsilon>0$ is a small constant added for numerical stability.
By conditioning $\alpha(\mathbf{c}),\beta(\mathbf{c})$ on the event representation $\mathbf{c}$, the model adapts timestep conditioned on heterogeneous event semantics, thereby modulating the denoising process and enabling event-specific restoration dynamics for the denoising targets.

\begin{algorithm}[t]
\caption{Sampling (Forecasting)}
\label{alg:validation_sampling}
\begin{algorithmic}[1]
\Require Initial state $\mathbf{Z}_0$; flow matching model $\mathcal{M}_\theta$;reparameterization function $\phi$; historical $p$ time series and event pairs $(\mathcal{H}_x, \mathcal{H}_c)$; evolving event sequence $\{\mathbf{c}_s\}_{s=p+1}^{q}=\mathcal{F}_\mathbf{c}$; Gaussian distribution $\mathcal{N}(0,\mathbf{I})$; total number of denoising steps $T$ and its incremental $\Delta_{\hat{t}}=\frac{1}{T}$.
\State Initialize forecasting set $\hat{\mathcal{X}} \gets \{\}$.
\For{$s = 1, \dots, p$}
    \State $(\mathbf{Z}_s, \_)=\mathcal{M}_\theta(\mathbf{Z}_{s-1}, \mathbf{x}_s, \mathbf{c}_s, t=1)$

\EndFor
\For{$s = p+1, \dots, q$}
    \State $\mathbf{n}_{s}^{0} \sim \mathcal{N}(0,\mathbf{I})$
    \For{$t = \Delta_{\hat{t}}, 2\Delta_{\hat{t}},\dots, 1$}
        \State $\Delta_{t} \gets \phi(t-\Delta_{\hat{t}}, \mathbf{c}_s)$
        \State $(\mathbf{Z}_s, \mathbf{\hat{v}}_{s}^{\Delta_{t}})=\mathcal{M}_\theta(\mathbf{Z}_{s-1}, \mathbf{n}_{s}^{\Delta_{t}}, \mathbf{c}_s, t)$
        \State $\mathbf{n}_{s}^{t}\gets\mathbf{n}_{s}^{\Delta_{t}} + \text{ODE}(\mathbf{\hat{v}}_{s}^{\Delta_{t}},\Delta_t)$
        \Comment{Refinement}

    \EndFor
    \State $\mathbf{\hat{x}}_{s} \gets \mathbf{n}_{s}^{1}$
    \State $\hat{\mathcal{X}} \gets \hat{\mathcal{X}} \cup \{\mathbf{\hat{x}}_s\}$
\EndFor
\State \Return $\hat{\mathcal{X}}$
\end{algorithmic}
\end{algorithm}

\section{Experiments}

\begin{table*}[!t]
\centering
\footnotesize
\renewcommand{\arraystretch}{1.08}
\setlength{\tabcolsep}{1.4mm}

\begin{tabular}{l c c c cc ccc c}
\toprule
\multirow{2}{*}{Method} & \multirow{2}{*}{Metric} & \multirow{2}{*}{Synthetic} & Atmosphere & \multicolumn{2}{c}{Traffic} & \multicolumn{3}{c}{Temperature--Rainfall} & \multirow{2}{*}{Avg.} \\
\cmidrule(lr){5-6} \cmidrule(lr){7-9}
 & & & Physics & Public & News & Houston & San~Fran. & New~York & \\
\midrule

\multirow{3}{*}{CSDI}
& CRPS & 0.5769 & 1.3840 & 0.4359 & 0.4471 & 0.5742 & 0.5986 & 0.8201 & 0.6910 \\
& WQL  & 0.3174 & 0.7452 & 0.2341 & 0.2445 & 0.3149 & 0.3274 & 0.4509 & 0.3763 \\
& MAE  & 0.8901 & 2.1477 & 0.7623 & 0.6585 & 0.8165 & 0.8905 & 1.1509 & 1.0452 \\
\midrule

\multirow{3}{*}{TimeDiff}
& CRPS & 0.6718 & 1.3876 & 0.6179 & 0.4977 & 0.8092 & 0.7194 & 1.0869 & 0.8272 \\
& WQL  & 0.3380 & 0.7475 & 0.3111 & 0.2509 & 0.4067 & 0.3618 & 0.5456 & 0.4231 \\
& MAE  & 0.6858 & 2.1343 & 0.6320 & 0.5114 & 0.8233 & 0.7335 & 1.1012 & 0.9459 \\
\midrule

\multirow{3}{*}{TMDM}
& CRPS & 0.1842 & 1.6422 & 0.2796 & 0.3059 & 0.5500 & 0.5532 & 0.6857 & 0.6001 \\
& WQL  & 0.0983 & 0.8233 & 0.1393 & 0.2012 & 0.2978 & 0.2987 & 0.3723 & 0.3187 \\
& MAE  & 0.2899 & 1.6568 & 0.3598 & 0.4480 & 0.7297 & 0.7305 & 0.9258 & 0.7344 \\
\midrule

\multirow{3}{*}{NsDiff}
& CRPS & 0.2101 & 1.2933 & 0.2825 & 0.4516 & 0.4924 & 0.6588 & 0.4585 & 0.5496 \\
& WQL  & 0.1131 & 0.6794 & 0.1406 & 0.3680 & 0.2769 & 0.2785 & 0.3612 & 0.3168 \\
& MAE  & 0.2657 & 1.5355 & 0.3508 & 0.6199 & 0.7015 & 0.6836 & 0.9451 & 0.7289 \\
\midrule

\multirow{3}{*}{DiffusionTS}
& CRPS & 0.6211 & 0.7151 & 0.5401 & 0.4437 & 0.6065 & 0.5453 & 0.8363 & 0.6154 \\
& WQL  & 0.3361 & 1.1631 & 0.2377 & 0.2388 & 0.3273 & 0.2938 & 0.4470 & 0.4348 \\
& MAE  & 0.8818 & 3.8218 & 0.5930 & 0.5988 & 0.8280 & 0.7314 & 1.1070 & 1.2231 \\
\midrule

\multirow{3}{*}{\textbf{\method}}
& CRPS & \textbf{0.0776} & \textbf{0.2857} & \textbf{0.1908} & \textbf{0.2943} & \textbf{0.2670} & \textbf{0.5183} & \textbf{0.2895} & \textbf{0.2747} \\
& WQL  & \textbf{0.0395} & \textbf{0.1464} & \textbf{0.1006} & \textbf{0.1561} & \textbf{0.1405} & \textbf{0.2702} & \textbf{0.1532} & \textbf{0.1438} \\
& MAE  & \textbf{0.0812} & \textbf{0.3544} & \textbf{0.2314} & \textbf{0.3561} & \textbf{0.3178} & \textbf{0.5869} & \textbf{0.3500} & \textbf{0.3254} \\
\bottomrule
\end{tabular}

\caption{Probabilistic forecasting results on seven event-aware datasets. We report CRPS$\downarrow$, WQL$\downarrow$, and MAE$\downarrow$ (lower is better) for each dataset, and the final column summarizes the average performance across datasets. Best results are highlighted in \textbf{bold}.}

\label{tab:comparison}
\end{table*}

We evaluate the performance of \method and address core research questions. Experiments are conducted on 7 datasets spanning 12 models. Configurations follow NsDiff~\cite{ye2025non} and standard TSF protocols~\cite{wu2023timesnet}. Results are averaged over three independent runs.
\begin{itemize}
    \item \textbf{RQ1}: How effectively does \method forecast non-stationary multimodal data in both deterministic and probabilistic settings?
    \item \textbf{RQ2}: When baselines can access event data, can they achieve performance comparable to \method?
    \item \textbf{RQ3}: How effective are the components, textual event conditioning and event-aware timestep in enhancing the \ method's performance?
    \item \textbf{RQ4}: How does \method's training efficiency compare to baselines and across forecasting horizons?
    \item \textbf{RQ5}: What insights can be gained from visualizations with vs. without events?
\end{itemize}
\subsection{Experimental Settings}
\noindent\textbf{Datasets}. The experiments are evaluated on one synthetic dataset and six real-world datasets spanning traffic, weather and atmospheric physics sector. These include Synthetic Dataset, Atmospheric Physics-Weather Events Dataset, Traffic–Public Events Dataset~\cite{liang2024exploring}, Temperature–Rainfall Events Dataset~\cite{lee2025timecap} and Traffic–News Events Dataset~\cite{wang2024news}. More Detailed descriptions are provided in the Appendix~\ref{datasetdescription}.

\noindent\textbf{Evaluation Metrics}.
We use comprehensive metrics for different forecasting tasks. For \emph{deterministic forecasting}, we use Mean Squared Error (MSE), Mean Absolute Error (MAE) and Root Mean Squared Error (RMSE) for evaluation.
For \emph{probabilistic forecasting},
we evaluate performance using Continuous Ranked Probability Score (CRPS) and Weighted Quantile Loss (WQL), and report MAE for point summaries.

\noindent\textbf{Baselines}. To evaluate \method's effectiveness, we benchmark it against probabilistic diffusion forecasting models and deterministic deep time series forecasting models.
For the probabilistic diffusion forecasting models, CSDI~\cite{tashiro2021csdi}, TimeDiff~\cite{shen2023non}, TMDM~\cite{li2024transformer}, DiffusionTS~\cite{yuan2024diffusion}, and NsDiff~\cite{ye2025non} are included. For the deterministic deep time series forecasting, we include representative models, including iTransformer~\cite{liu2023itransformer}, Koopa~\cite{liu2023koopa}, NSTransformer~\cite{liu2022non}, PatchTST~\cite{nie2022time}, TimesNet~\cite{wu2023timesnet}, and other used baselines. NsDiff~\cite{ye2025non}, Koopa~\cite{liu2023koopa}, and NSTransformer~\cite{liu2022non} are designed for the non-stationary time series forecasting.


\begin{table*}[ht]
\centering
\footnotesize
\renewcommand{\arraystretch}{1.08}
\setlength{\tabcolsep}{1.4mm}

\begin{tabular}{l c c c cc ccc c}
\toprule
\multirow{2}{*}{Method} & \multirow{2}{*}{Metric} & \multirow{2}{*}{Synthetic} & Atmosphere & \multicolumn{2}{c}{Traffic} & \multicolumn{3}{c}{Temperature--Rainfall} & \multirow{2}{*}{Avg.} \\
\cmidrule(lr){5-6} \cmidrule(lr){7-9}
 & & & Physics & Public & News & Houston & San~Fran. & New~York & \\
\midrule

\multirow{3}{*}{Autoformer}
& MAE  & 0.3141 & 2.0739 & 0.4296 & 0.3536 & 0.7127 & 0.7011 & 0.9930 & 0.7969 \\
& MSE  & 0.2612 & 6.5844 & 0.5258 & 0.2144 & 0.8194 & 0.8168 & 1.5016 & 1.5319 \\
& RMSE & 0.5161 & 2.5656 & 0.7294 & 0.4673 & 0.9062 & 0.9043 & 1.2215 & 1.0443 \\
\midrule

\multirow{3}{*}{DLinear}
& MAE  & 0.3614 & 1.4269 & 0.4262 & 0.3165 & 0.7277 & 0.6016 & 0.9305 & 0.6844 \\
& MSE  & 0.3022 & 3.5741 & 0.5786 & 0.1641 & 0.7980 & 0.5992 & 1.1781 & 1.0278 \\
& RMSE & 0.5493 & 1.8910 & 0.7617 & 0.4054 & 0.8943 & 0.7748 & 1.0852 & 0.9088 \\
\midrule

\multirow{3}{*}{iTransformer}
& MAE  & 0.3821 & 1.5398 & 0.4419 & 0.3136 & 0.7239 & 0.6275 & 0.9420 & 0.7101 \\
& MSE  & 0.3244 & 4.2576 & 0.6249 & 0.1657 & 0.8524 & 0.6970 & 1.3635 & 1.1836 \\
& RMSE & 0.5705 & 2.0634 & 0.7909 & 0.4067 & 0.9231 & 0.8358 & 1.1689 & 0.9656 \\
\midrule

\multirow{3}{*}{Koopa}
& MAE  & 0.3233 & 1.6498 & 0.4161 & \textbf{0.3057} & 0.7226 & 0.6189 & 0.8582 & 0.6992 \\
& MSE  & 0.2680 & 5.2355 & 0.5285 & \textbf{0.1640} & 0.8509 & 0.6462 & 1.1296 & 1.2604 \\
& RMSE & 0.5188 & 2.2884 & 0.7271 & \textbf{0.4054} & 0.9227 & 0.7852 & 1.0635 & 0.9587 \\
\midrule

\multirow{3}{*}{NSTransformer}
& MAE  & 0.3261 & 1.7127 & 0.4569 & 0.3626 & 0.7698 & 0.6878 & 0.9752 & 0.7559 \\
& MSE  & 0.2762 & 5.7245 & 0.6027 & 0.2198 & 0.9331 & 0.7514 & 1.3522 & 1.4086 \\
& RMSE & 0.5256 & 2.3902 & 0.7763 & 0.4672 & 0.9695 & 0.8677 & 1.1627 & 1.0227 \\
\midrule

\multirow{3}{*}{PatchTST}
& MAE  & 0.3228 & 1.8450 & 0.4029 & 0.3371 & 0.7743 & 0.6456 & 0.9150 & 0.7490 \\
& MSE  & 0.2614 & 5.7711 & 0.4693 & 0.1985 & 0.9516 & 0.7223 & 1.2445 & 1.3741 \\
& RMSE & 0.5118 & 2.4012 & 0.6890 & 0.4457 & 0.9750 & 0.8492 & 1.1135 & 0.9979 \\
\midrule

\multirow{3}{*}{TimesNet}
& MAE  & 0.3181 & 1.5708 & 0.4283 & 0.3263 & 0.7159 & 0.6367 & 0.9269 & 0.7033 \\
& MSE  & 0.2633 & 4.4105 & 0.5914 & 0.1861 & 0.8446 & 0.7087 & 1.3148 & 1.1885 \\
& RMSE & 0.5132 & 2.1030 & 0.7699 & 0.4321 & 0.9187 & 0.8462 & 1.1496 & 0.9618 \\
\midrule

\multirow{3}{*}{\textbf{\method}}
& MAE  & \textbf{0.0812} & \textbf{0.3544} & \textbf{0.2314} & 0.3561 & \textbf{0.3500} & \textbf{0.3178} & \textbf{0.5869} & \textbf{0.3254} \\
& MSE  & \textbf{0.0235} & \textbf{0.3668} & \textbf{0.2243} & 0.3936 & \textbf{0.3512} & \textbf{0.3202} & \textbf{1.1526} & \textbf{0.4046} \\
& RMSE & \textbf{0.1535} & \textbf{0.6056} & \textbf{0.4736} & 0.6273 & \textbf{0.5926} & \textbf{0.5659} & \textbf{1.0736} & \textbf{0.5846} \\
\bottomrule
\end{tabular}

\caption{Deterministic forecasting results on seven event-aware datasets. We report MAE$\downarrow$, MSE$\downarrow$, and RMSE$\downarrow$ (lower is better) for each dataset, and the last column summarizes the average performance across datasets. Best results are highlighted in \textbf{bold}.}

\label{tab:deterministic-comparison}
\end{table*}

\subsection{Performance Comparison on Probabilistic and Deterministic Forecasting (\texorpdfstring{\underline{RQ1}}{RQ1})}
\method achieves superior performance across both deterministic and probabilistic forecasting tasks.
Table~\ref{tab:comparison} compares probabilistic forecasting performance across seven datasets using CRPS, WQL, and average MAE, where lower values indicate better performance.
The results highlight \method's superior performance across diverse event-aware forecasting settings, with average improvements of 41.3\% and 27.5\% across all datasets and metrics for probabilistic and deterministic forecasting, respectively.
Specifically, \method delivers notable gains, lowering the CRPS on the Synthetic dataset from TMDM's $0.1842$ to $0.0776$, a 58.4\% improvement, and reducing the CRPS on the Atmosphere Physics dataset from DiffusionTS's $0.7151$ to $0.2857$, a 65.1\% improvement.
Table~\ref{tab:deterministic-comparison} reports deterministic forecasting results across datasets.
On the Synthetic dataset, \method achieves an MAE of $0.0812$, an $74.1\%$ reduction from Autoformer's $0.3141$; on the averaged dataset results, \method achieves an RMSE of $0.5846$, denoting a $35.8\%$ reduction from DLinear's $0.9088$.
Deterministic forecasting models perform competitively on the Traffic–News dataset, where the limited contextual relevance of long-form, low-information textual event representations significantly weakens the effectiveness of event-aware forecasting, as further discussed in Appendix~\ref{datasetdescription} and Appendix~\ref{datasetanalysis}.

\begin{figure}[t]
    \centering
    \includegraphics[width=1\linewidth]{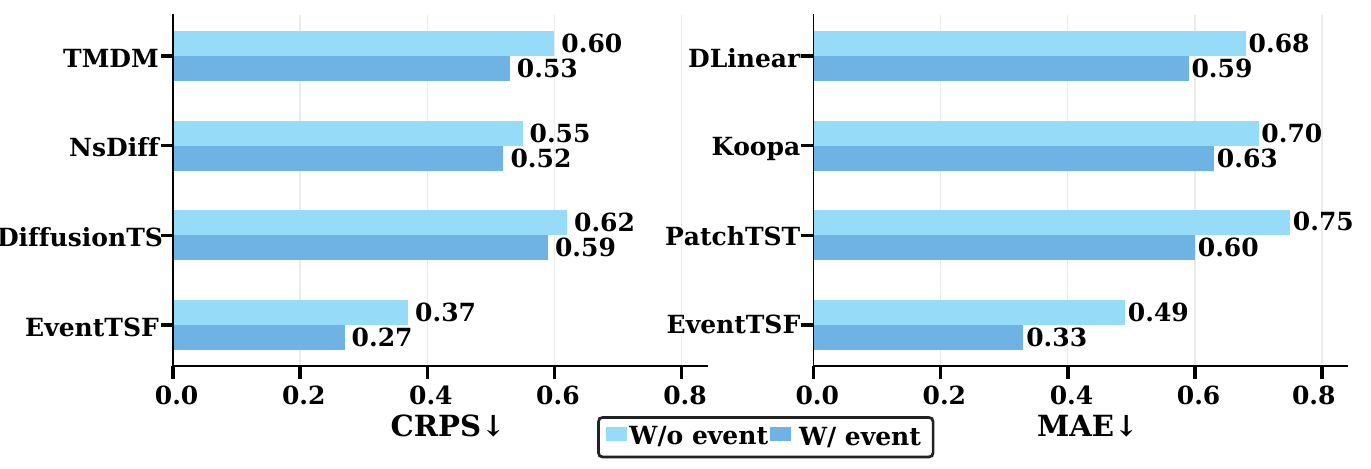}
    \caption{Baselines with Event Insert. (\textbf{Left}) Probabilistic forecasting(CRPS$\downarrow$); (\textbf{Right})Deterministic forecasting (MAE$\downarrow$)}
    \label{fig:addevent}
\end{figure}

\begin{figure}[t]
    \centering
    \includegraphics[width=0.95\linewidth]{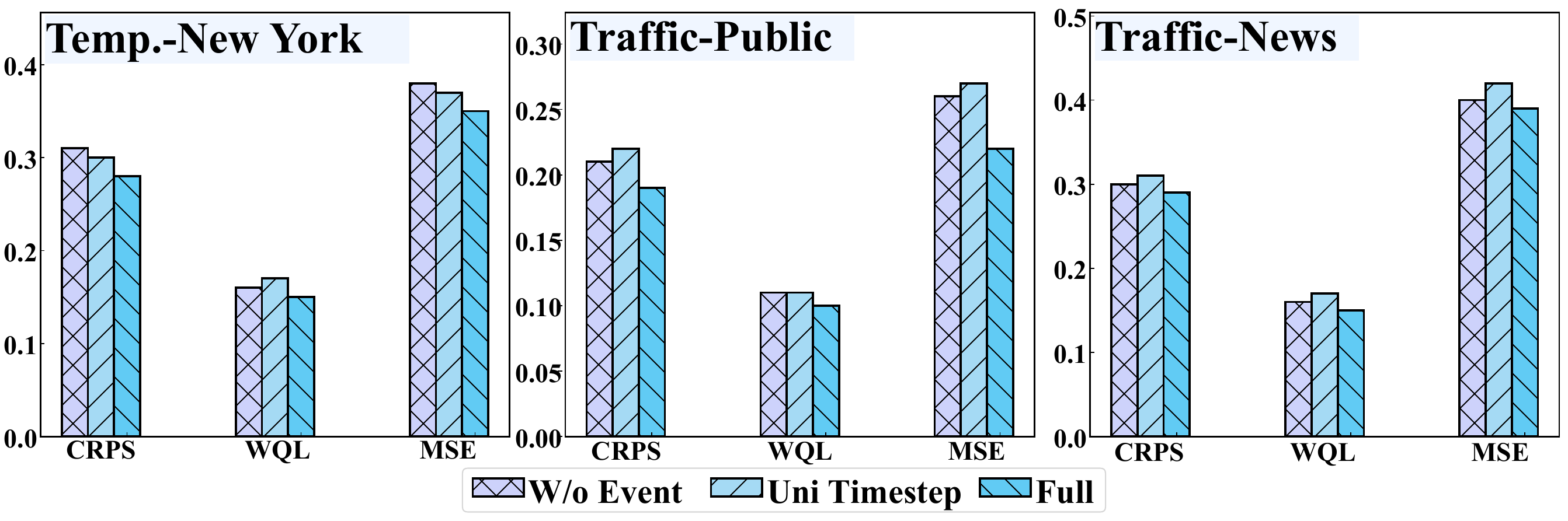}
    \caption{Ablation. Evaluating textual events and event-aware diffusion timestep against the full model.
    }
    \label{fig:ablation_study}
\end{figure}

\subsection{Event-Enhanced Baseline Performance (\texorpdfstring{\underline{RQ2}}{RQ2})}
Most existing baselines cannot natively incorporate external textual events; therefore, we augment both probabilistic and deterministic baselines with the aligned event embeddings for a fairer evaluation. For probabilistic forecasting baselines, we replace the denoiser’s transformer block with the crossformer block to enable event-conditioned cross-attention.
For deterministic forecasting baselines, we linearly project event embeddings into input embedding space and add them to the input tokens as a residual conditioning signal.

As reported in Fig.~\ref{fig:addevent}, incorporating events consistently improves the performance of both probabilistic and deterministic baselines.
Even with event access, all baselines remain substantially inferior to \method.
For probabilistic forecasting, incorporating event information yields a CRPS reduction of $0.07$ for TMDM, corresponding to an $11.7\%$ relative improvement. In comparison, \method achieves a larger CRPS reduction of $0.10$, translating to a $27.1\%$ improvement. For deterministic forecasting, under the MAE metric, \method with textual event conditioning consistently outperforms the other event-augmented baselines, delivering an average gain of $0.28$ corresponding to an $45.9\%$ relative improvement.
Overall, even with the textual event access, none of the event-enhanced baselines achieves performance comparable to \method.
This indicates that injecting event embeddings simply is insufficient to exploiting event information and achieve event-aware modeling.

\subsection{Ablation Study (\texorpdfstring{\underline{RQ3}}{RQ3})}
Figure~\ref{fig:ablation_study} reports ablation results across three datasets under three experiment configurations and evaluated with three metrics.
Removing textual events causes performance degradation: WQL increases $10.71\%$ on the Temperature–New~York events dataset. Moreover, replacing classical uniform diffusion timestep sampling with event-aware diffusion timestep yields consistent gains, reducing CRPS by $9.21\%$, WQL by $12.50\%$, and MSE by $10.42\%$. Additional detailed experimental on the effectiveness of event-aware diffusion timestep sampling are provided in Appendix~\ref{app:eventawaredifferenttimestep}. Overall, the results consistently confirm the necessity of each component, demonstrating that the complete architecture is essential for event-aware non-stationary time series forecasting.

\subsection{Efficiency Analysis (\texorpdfstring{\underline{RQ4}}{RQ4})}
\begin{figure}[t]
    \centering
    \includegraphics[width=1\linewidth]{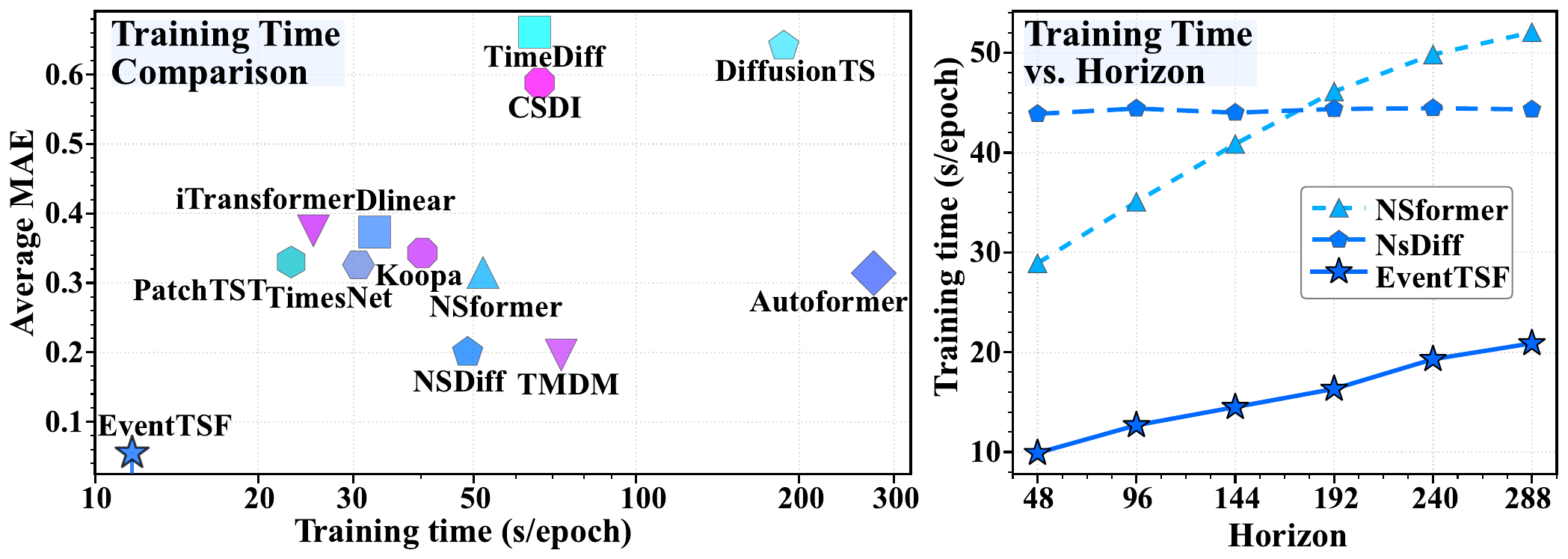}
    \caption{Training Efficiency. \textbf{(Left)} Training time vs. MAE. \textbf{(Right)} Training time scalability with horizon length.}
    \label{fig:efficiency}
\end{figure}

Figure~\ref{fig:efficiency} analyzes training efficiency across forecasting models and scalability with horizon lengths. \method achieves the shortest training time and lowest forecasting error across all datasets. It shows $1.13\times$ faster training and $4.85\times$ lower forecasting error versus the second-best baseline on the Synthetic dataset.
NSTransformer shows linear training time growth with horizon length, while NsDiff maintains consistently high training time across all horizons. In contrast, \method achieves both lower training time and sublinear growth with increasing horizons. Because it generates forecasts segment-wise rather than point-wise.
Results confirm \ method's superior efficiency in event-aware non-stationary forecasting, benefiting long-horizon scenarios.

\subsection{Event-Aware Forecasting Visualization (\texorpdfstring{\underline{RQ5}}{RQ5})}

\begin{figure}[t]
    \centering
    \includegraphics[width=1\linewidth]{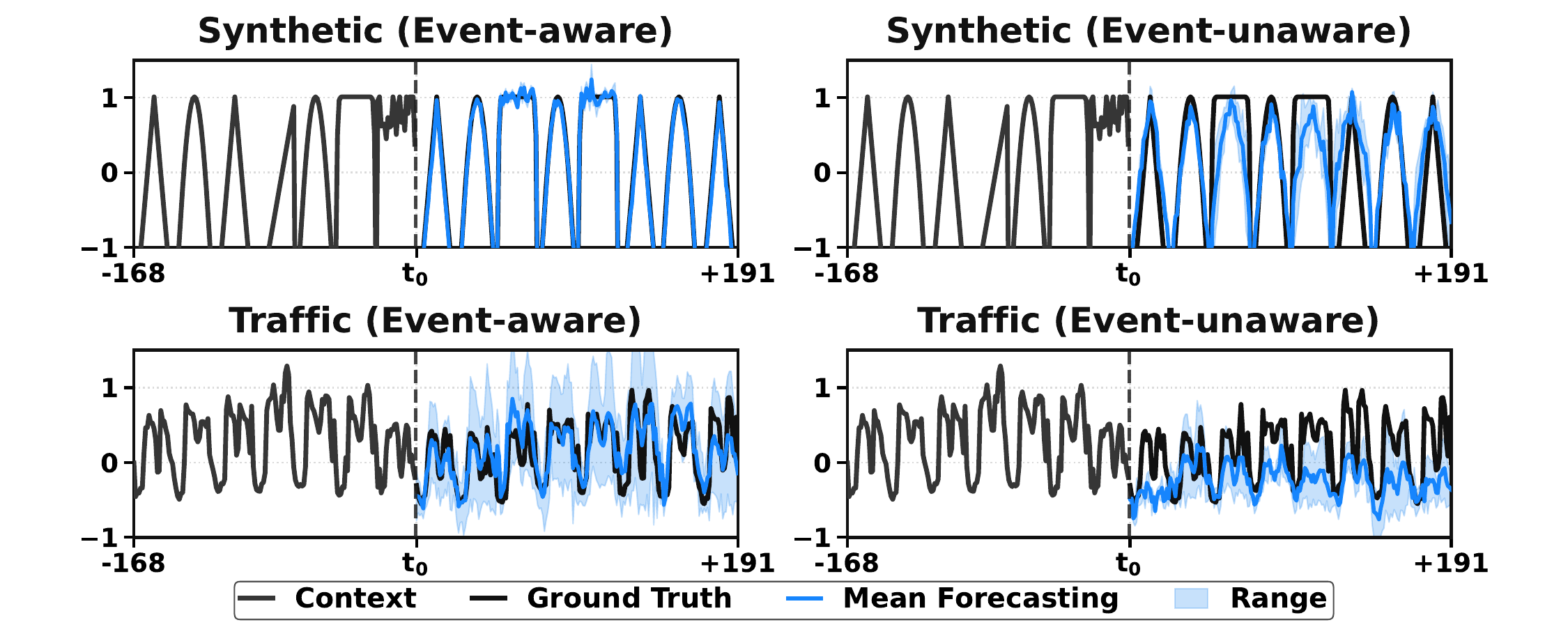}
    \caption{Visualization of event-aware forecasting on the Synthetic and Traffic–Public datasets. External events improve accuracy.}
    \label{fig:visualization}
\end{figure}

Figure~\ref{fig:visualization} illustrates event-aware forecasting visualization on the Synthetic and Traffic-Public datasets.
Textual event incorporation significantly improves accuracy. On the Synthetic dataset, the event-aware model accurately captures abrupt transitions while the event-unaware model produces smoothed responses. On the Traffic-Public dataset, event information improves the alignment with ground truth under noisy conditions, although improvements are more pronounced on the synthetic data.
Results demonstrate that event information provides essential semantic cues for precise forecasting across both synthetic dataset and real-world dataset.


\section{Related Work}
\subsection{Non-stationary Time Series Forecasting}
Non-stationarity challenges time series forecasting; deep learning tackles it through normalization, model design, and learning theory.
\textbf{(I) Normalization plugins} are designed from time domains~\cite{kim2021reversible,fan2023dish} and frequency domains~\cite{ye2024frequency,liutimestacker}.
\textbf{(II) Models} include tailored architectures~\cite{liu2024timebridge,liu2022non}, such as frequency decompositions and statistical decompositions~\cite{yi2023frequency,luo2025tfdnet,ge2024moment}, and dynamic system modeling~\cite{liu2023koopa,wang2022koopman}.
\textbf{(III) Learning-theoretic approaches} handle distribution shifts via domain generation~\cite{liu2024time}, adaptation~\cite{kim2025battling,li2022ddg}, and online learning~\cite{wen2023onenet,zhan2025continuous}.
However, incorporating external modality for non-stationary time series forecasting remains largely unexplored.

\subsection{Multimodal for Time series Forecasting}
Incorporating textual information into time series forecasting has gained attention across finance, traffic, and power systems~\cite{xu2021rest,liang2024exploring,lian2026contextual,bai2024news}, showing promise in early LLM-based forecasting~\cite{xue2023promptcast,jin2024position}. Subsequently,verious fusion methods have emerged exploring model architecture~\cite{xu2024beyond}, learning strategies~\cite{liu2025calf}, and benchmarks~\cite{williams2024context}.
Recently, retrieval-augmented~\cite{jiang2025explainable,zhang2025timeraf}, reasoning~\cite{guan2025timeomni}, and agent-based systems~\cite{lee2025timecap,wang2024news} leverage LLMs' contextual analysis capabilities. However, these approaches rely heavily on LLMs for modality fusion, potentially limiting their generative expressiveness.
Autoregressive diffusion architectures offer probabilistic forecasting capabilities while supporting textual conditioning. Combining autoregressive mechanisms with diffusion modeling remains unexplored for event-aware time series forecasting.


\section{Conclusion}
This work pioneers the integration of external multimodal knowledge to tackle non-stationarity in time series forecasting. \method captures fine-grained time series and event interactions via an autoregressive diffusion architecture with event-aware timestep.
This novel architecture addresses the gap in synchronizing and modeling discrete event and continuous time series.
Event-aware timestep reallocate timesteps conditioned on the event so that the denoiser sees a more balanced difficulty distribution across diffusion stages.
Extensive experiments on 12 baselines show its effectiveness and highlight \method as a flexible paradigm for incorporating heterogeneous event semantics into temporal modeling.
Success in modeling cross-modal non-stationary dynamics opens up a direction for non-stationary time series analysis.

\clearpage

\bibliographystyle{named}
\bibliography{ref_main,ref_app}

@article{aliang2024exploringa,
  title={Exploring large language models for human mobility prediction under public events},
  author={Liang, Yuebing and Liu, Yichao and Wang, Xiaohan and Zhao, Zhan},
  journal={Computers, Environment and Urban Systems},
  volume={112},
  pages={102153},
  year={2024},
  publisher={Elsevier}
}

@inproceedings{alee2025timecapa,
  title={Timecap: Learning to contextualize, augment, and predict time series events with large language model agents},
  author={Lee, Geon and Yu, Wenchao and Shin, Kijung and Cheng, Wei and Chen, Haifeng},
  booktitle={Proceedings of the AAAI Conference on Artificial Intelligence},
  volume={39},
  pages={18082--18090},
  year={2025}
}

@article{awang2024newsa,
  title={From news to forecast: Integrating event analysis in llm-based time series forecasting with reflection},
  author={Wang, Xinlei and Feng, Maike and Qiu, Jing and Gu, Jinjin and Zhao, Junhua},
  journal={Advances in Neural Information Processing Systems},
  volume={37},
  pages={58118--58153},
  year={2024}
}

@article{anarasimhan2024timea,
  title={Time weaver: A conditional time series generation model},
  author={Narasimhan, Sai Shankar and Agarwal, Shubhankar and Akcin, Oguzhan and Sanghavi, Sujay and Chinchali, Sandeep},
  journal={arXiv preprint arXiv:2403.02682},
  year={2024}
}

@inproceedings{aronneberger2015ua,
  title={U-net: Convolutional networks for biomedical image segmentation},
  author={Ronneberger, Olaf and Fischer, Philipp and Brox, Thomas},
  booktitle={International Conference on Medical image computing and computer-assisted intervention},
  pages={234--241},
  year={2015},
  organization={Springer}
}

@inproceedings{aliu2018hierarchicala,
  title={Hierarchical complementary attention network for predicting stock price movements with news},
  author={Liu, Qikai and Cheng, Xiang and Su, Sen and Zhu, Shuguang},
  booktitle={Proceedings of the 27th ACM International Conference on Information and Knowledge Management},
  pages={1603--1606},
  year={2018}
}

@inproceedings{axu2021resta,
  title={Rest: Relational event-driven stock trend forecasting},
  author={Xu, Wentao and Liu, Weiqing and Xu, Chang and Bian, Jiang and Yin, Jian and Liu, Tie-Yan},
  booktitle={Proceedings of the web conference 2021},
  pages={1--10},
  year={2021}
}

@article{aemami2023modalitya,
  title={Modality-aware Transformer for Financial Time series Forecasting},
  author={Emami, Hajar and Dang, Xuan-Hong and Shah, Yousaf and Zerfos, Petros},
  journal={arXiv preprint arXiv:2310.01232},
  year={2023}
}

@inproceedings{awang2021deeptradera,
  title={Deeptrader: a deep reinforcement learning approach for risk-return balanced portfolio management with market conditions embedding},
  author={Wang, Zhicheng and Huang, Biwei and Tu, Shikui and Zhang, Kun and Xu, Lei},
  booktitle={Proceedings of the AAAI conference on artificial intelligence},
  volume={35},
  pages={643--650},
  year={2021}
}

@inproceedings{aduan2022factorvaea,
  title={Factorvae: A probabilistic dynamic factor model based on variational autoencoder for predicting cross-sectional stock returns},
  author={Duan, Yitong and Wang, Lei and Zhang, Qizhong and Li, Jian},
  booktitle={Proceedings of the AAAI conference on artificial intelligence},
  volume={36},
  pages={4468--4476},
  year={2022}
}

@inproceedings{axia2024marketa,
  title={Market-gan: Adding control to financial market data generation with semantic context},
  author={Xia, Haochong and Sun, Shuo and Wang, Xinrun and An, Bo},
  booktitle={Proceedings of the AAAI Conference on Artificial Intelligence},
  volume={38},
  pages={15996--16004},
  year={2024}
}

@article{anie2024surveya,
  title={A survey of large language models for financial applications: Progress, prospects and challenges},
  author={Nie, Yuqi and Kong, Yaxuan and Dong, Xiaowen and Mulvey, John M and Poor, H Vincent and Wen, Qingsong and Zohren, Stefan},
  journal={arXiv preprint arXiv:2406.11903},
  year={2024}
}

@article{aliu2025llm4ftsa,
  title={LLM4FTS: Enhancing Large Language Models for Financial Time Series Prediction},
  author={Liu, Zian and Jia, Renjun},
  journal={arXiv preprint arXiv:2505.02880},
  year={2025}
}

@article{axiao2025retrievala,
  title={Retrieval-augmented large language models for financial time series forecasting},
  author={Xiao, Mengxi and Jiang, Zihao and Qian, Lingfei and Chen, Zhengyu and He, Yueru and Xu, Yijing and Jiang, Yuecheng and Li, Dong and Weng, Ruey-Ling and Peng, Min and others},
  journal={arXiv preprint arXiv:2502.05878},
  year={2025}
}

@article{ali2024marsa,
  title={Mars: a financial market simulation engine powered by generative foundation model},
  author={Li, Junjie and Liu, Yang and Liu, Weiqing and Fang, Shikai and Wang, Lewen and Xu, Chang and Bian, Jiang},
  journal={arXiv preprint arXiv:2409.07486},
  year={2024}
}

@article{azhang2025camefa,
  title={CAMEF: Causal-Augmented Multi-Modality Event-Driven Financial Forecasting by Integrating Time Series Patterns and Salient Macroeconomic Announcements},
  author={Zhang, Yang and Yang, Wenbo and Wang, Jun and Ma, Qiang and Xiong, Jie},
  journal={arXiv preprint arXiv:2502.04592},
  year={2025}
}

@article{ayang2025learninga,
  title={Learning Universal Multi-level Market Irrationality Factors to Improve Stock Return Forecasting},
  author={Yang, Chen and Wang, Jingyuan and Jiang, Xiaohan and Wu, Junjie},
  journal={arXiv preprint arXiv:2502.04737},
  year={2025}
}

@article{yuan2024diffusion,
  title={Diffusion-ts: Interpretable diffusion for general time series generation},
  author={Yuan, Xinyu and Qiao, Yan},
  journal={arXiv preprint arXiv:2403.01742},
  year={2024}
}

@article{chen2024diffusion,
  title={Diffusion forcing: Next-token prediction meets full-sequence diffusion},
  author={Chen, Boyuan and Mart{\'\i} Mons{\'o}, Diego and Du, Yilun and Simchowitz, Max and Tedrake, Russ and Sitzmann, Vincent},
  journal={Advances in Neural Information Processing Systems},
  volume={37},
  pages={24081--24125},
  year={2024}
}

@article{ye2025non,
  title={Non-stationary Diffusion For Probabilistic Time Series Forecasting},
  author={Ye, Weiwei and Xu, Zhuopeng and Gui, Ning},
  journal={arXiv preprint arXiv:2505.04278},
  year={2025}
}

@article{tashiro2021csdi,
  title={Csdi: Conditional score-based diffusion models for probabilistic time series imputation},
  author={Tashiro, Yusuke and Song, Jiaming and Song, Yang and Ermon, Stefano},
  journal={Advances in neural information processing systems},
  volume={34},
  pages={24804--24816},
  year={2021}
}

@inproceedings{shen2023non,
  title={Non-autoregressive conditional diffusion models for time series prediction},
  author={Shen, Lifeng and Kwok, James},
  booktitle={ICML},
  pages={31016--31029},
  year={2023},
  organization={PMLR}
}

@inproceedings{li2024transformer,
  title={Transformer-modulated diffusion models for probabilistic multivariate time series forecasting},
  author={Li, Yuxin and Chen, Wenchao and Hu, Xinyue and Chen, Bo and Zhou, Mingyuan and others},
  booktitle={ICLR},
  year={2024}
}

@inproceedings{wu2023timesnet,
  title={TimesNet: Temporal 2D-Variation Modeling for General Time Series Analysis},
  author={Haixu Wu and Tengge Hu and Yong Liu and Hang Zhou and Jianmin Wang and Mingsheng Long},
  booktitle={ICLR},
  year={2023}
}

@article{liu2023itransformer,
  title={itransformer: Inverted transformers are effective for time series forecasting},
  author={Liu, Yong and Hu, Tengge and Zhang, Haoran and Wu, Haixu and Wang, Shiyu and Ma, Lintao and Long, Mingsheng},
  journal={arXiv preprint arXiv:2310.06625},
  year={2023}
}

@article{liu2023koopa,
  title={Koopa: Learning non-stationary time series dynamics with koopman predictors},
  author={Liu, Yong and Li, Chenyu and Wang, Jianmin and Long, Mingsheng},
  journal={Advances in neural information processing systems},
  volume={36},
  pages={12271--12290},
  year={2023}
}

@article{nie2022time,
  title={A time series is worth 64 words: Long-term forecasting with transformers},
  author={Nie, Yuqi and Nguyen, Nam H and Sinthong, Phanwadee and Kalagnanam, Jayant},
  journal={arXiv preprint arXiv:2211.14730},
  year={2022}
}

@inproceedings{kim2025battling,
  title={Battling the non-stationarity in time series forecasting via test-time adaptation},
  author={Kim, HyunGi and Kim, Siwon and Mok, Jisoo and Yoon, Sungroh},
  booktitle={AAAI},
  volume={39},
  pages={17868--17876},
  year={2025}
}

@article{liu2024timebridge,
  title={Timebridge: Non-stationarity matters for long-term time series forecasting},
  author={Liu, Peiyuan and Wu, Beiliang and Hu, Yifan and Li, Naiqi and Dai, Tao and Bao, Jigang and Xia, Shu-tao},
  journal={arXiv preprint arXiv:2410.04442},
  year={2024}
}

@article{liu2022non,
  title={Non-stationary transformers: Exploring the stationarity in time series forecasting},
  author={Liu, Yong and Wu, Haixu and Wang, Jianmin and Long, Mingsheng},
  journal={Advances in neural information processing systems},
  volume={35},
  pages={9881--9893},
  year={2022}
}

@article{luo2025tfdnet,
  title={TFDNet: Time--Frequency enhanced Decomposed Network for long-term time series forecasting},
  author={Luo, Yuxiao and Zhang, Songming and Lyu, Ziyu and Hu, Yuhan},
  journal={Pattern Recognition},
  volume={162},
  pages={111412},
  year={2025},
  publisher={Elsevier}
}

@inproceedings{kim2021reversible,
  title={Reversible instance normalization for accurate time-series forecasting against distribution shift},
  author={Kim, Taesung and Kim, Jinhee and Tae, Yunwon and Park, Cheonbok and Choi, Jang-Ho and Choo, Jaegul},
  booktitle={ICLR},
  year={2021}
}

@inproceedings{fan2023dish,
  title={Dish-ts: a general paradigm for alleviating distribution shift in time series forecasting},
  author={Fan, Wei and Wang, Pengyang and Wang, Dongkun and Wang, Dongjie and Zhou, Yuanchun and Fu, Yanjie},
  booktitle={AAAI},
  volume={37},
  pages={7522--7529},
  year={2023}
}

@article{ye2024frequency,
  title={Frequency adaptive normalization for non-stationary time series forecasting},
  author={Ye, Weiwei and Deng, Songgaojun and Zou, Qiaosha and Gui, Ning},
  journal={Advances in Neural Information Processing Systems},
  volume={37},
  pages={31350--31379},
  year={2024}
}

@article{wang2022koopman,
  title={Koopman neural forecaster for time series with temporal distribution shifts},
  author={Wang, Rui and Dong, Yihe and Arik, Sercan {\"O} and Yu, Rose},
  journal={arXiv preprint arXiv:2210.03675},
  year={2022}
}

@article{yi2023frequency,
  title={Frequency-domain MLPs are more effective learners in time series forecasting},
  author={Yi, Kun and Zhang, Qi and Fan, Wei and Wang, Shoujin and Wang, Pengyang and He, Hui and An, Ning and Lian, Defu and Cao, Longbing and Niu, Zhendong},
  journal={Advances in Neural Information Processing Systems},
  volume={36},
  pages={76656--76679},
  year={2023}
}

@inproceedings{li2022ddg,
  title={Ddg-da: Data distribution generation for predictable concept drift adaptation},
  author={Li, Wendi and Yang, Xiao and Liu, Weiqing and Xia, Yingce and Bian, Jiang},
  booktitle={AAAI},
  volume={36},
  pages={4092--4100},
  year={2022}
}

@article{zhan2025continuous,
  title={Continuous Evolution Pool: Taming Recurring Concept Drift in Online Time Series Forecasting},
  author={Zhan, Tianxiang and Jin, Ming and He, Yuanpeng and Liang, Yuxuan and Deng, Yong and Pan, Shirui},
  journal={arXiv preprint arXiv:2506.14790},
  year={2025}
}

@article{lipman2022flow,
  title={Flow matching for generative modeling},
  author={Lipman, Yaron and Chen, Ricky TQ and Ben-Hamu, Heli and Nickel, Maximilian and Le, Matt},
  journal={arXiv preprint arXiv:2210.02747},
  year={2022}
}

@article{bai2024news,
  title={News and load: A quantitative exploration of natural language processing applications for forecasting day-ahead electricity system demand},
  author={Bai, Yun and Camal, Simon and Michiorri, Andrea},
  journal={IEEE Transactions on Power Systems},
  volume={39},
  pages={6222--6234},
  year={2024},
  publisher={IEEE}
}

@article{xue2023promptcast,
  title={Promptcast: A new prompt-based learning paradigm for time series forecasting},
  author={Xue, Hao and Salim, Flora D},
  journal={IEEE Transactions on Knowledge and Data Engineering},
  volume={36},
  pages={6851--6864},
  year={2023},
  publisher={IEEE}
}

@inproceedings{jin2024position,
  title={Position: What can large language models tell us about time series analysis},
  author={Jin, Ming and Zhang, Yifan and Chen, Wei and Zhang, Kexin and Liang, Yuxuan and Yang, Bin and Wang, Jindong and Pan, Shirui and Wen, Qingsong},
  booktitle={ICML},
  year={2024},
  organization={MLResearchPress}
}

@inproceedings{liu2025calf,
  title={Calf: Aligning llms for time series forecasting via cross-modal fine-tuning},
  author={Liu, Peiyuan and Guo, Hang and Dai, Tao and Li, Naiqi and Bao, Jigang and Ren, Xudong and Jiang, Yong and Xia, Shu-Tao},
  booktitle={AAAI},
  volume={39},
  pages={18915--18923},
  year={2025}
}

@article{xu2024beyond,
  title={Beyond trend and periodicity: Guiding time series forecasting with textual cues},
  author={Xu, Zhijian and Bian, Yuxuan and Zhong, Jianyuan and Wen, Xiangyu and Xu, Qiang},
  journal={arXiv e-prints},
  pages={arXiv--2405},
  year={2024}
}

@article{jiang2025explainable,
  title={Explainable multi-modal time series prediction with llm-in-the-loop},
  author={Jiang, Yushan and Yu, Wenchao and Lee, Geon and Song, Dongjin and Shin, Kijung and Cheng, Wei and Liu, Yanchi and Chen, Haifeng},
  journal={arXiv preprint arXiv:2503.01013},
  year={2025}
}

@article{zhang2025timeraf,
  title={Timeraf: Retrieval-augmented foundation model for zero-shot time series forecasting},
  author={Zhang, Huanyu and Xu, Chang and Zhang, Yi-Fan and Zhang, Zhang and Wang, Liang and Bian, Jiang},
  journal={IEEE Transactions on Knowledge and Data Engineering},
  year={2025},
  publisher={IEEE}
}

@article{wen2023onenet,
  title={Onenet: Enhancing time series forecasting models under concept drift by online ensembling},
  author={Wen, Qingsong and Chen, Weiqi and Sun, Liang and Zhang, Zhang and Wang, Liang and Jin, Rong and Tan, Tieniu and others},
  journal={Advances in Neural Information Processing Systems},
  volume={36},
  pages={69949--69980},
  year={2023}
}

@inproceedings{liutimestacker,
  title={TimeStacker: A Novel Framework with Multilevel Observation for Capturing Nonstationary Patterns in Time Series Forecasting},
  author={Liu, Qinglong and Xu, Cong and Jiang, Wenhao and Wang, Kaixuan and Ma, Lin and Li, Haifeng},
  booktitle={ICML},
  year={2025}
}

@article{liu2022flow,
  title={Flow straight and fast: Learning to generate and transfer data with rectified flow},
  author={Liu, Xingchao and Gong, Chengyue and Liu, Qiang},
  journal={arXiv preprint arXiv:2209.03003},
  year={2022}
}

@article{williams2024context,
  title={Context is key: A benchmark for forecasting with essential textual information},
  author={Williams, Andrew Robert and Ashok, Arjun and Marcotte, {\'E}tienne and Zantedeschi, Valentina and Subramanian, Jithendaraa and Riachi, Roland and Requeima, James and Lacoste, Alexandre and Rish, Irina and Chapados, Nicolas and others},
  journal={arXiv preprint arXiv:2410.18959},
  year={2024}
}

@article{zhang2025does,
  title={Does Multimodality Lead to Better Time Series Forecasting?},
  author={Zhang, Xiyuan and Han, Boran and Fang, Haoyang and Ansari, Abdul Fatir and Zhang, Shuai and Maddix, Danielle C and Hu, Cuixiong and Wilson, Andrew Gordon and Mahoney, Michael W and Wang, Hao and others},
  journal={arXiv preprint arXiv:2506.21611},
  year={2025}
}

@inproceedings{liang2024foundation,
  title={Foundation models for time series analysis: A tutorial and survey},
  author={Liang, Yuxuan and Wen, Haomin and Nie, Yuqi and Jiang, Yushan and Jin, Ming and Song, Dongjin and Pan, Shirui and Wen, Qingsong},
  booktitle={KDD},
  pages={6555--6565},
  year={2024}
}

@article{box1976time,
  title={Time series analysis. Forecasting and control},
  author={Box, George EP and Jenkins, Gwilym M},
  journal={Holden-Day Series in Time Series Analysis},
  year={1976}
}

@article{liu2024time,
  title={Time-series forecasting for out-of-distribution generalization using invariant learning},
  author={Liu, Haoxin and Kamarthi, Harshavardhan and Kong, Lingkai and Zhao, Zhiyuan and Zhang, Chao and Prakash, B Aditya},
  journal={arXiv preprint arXiv:2406.09130},
  year={2024}
}

@inproceedings{choi2022perception,
  title={Perception prioritized training of diffusion models},
  author={Choi, Jooyoung and Lee, Jungbeom and Shin, Chaehun and Kim, Sungwon and Kim, Hyunwoo and Yoon, Sungroh},
  booktitle={CVPR},
  pages={11472--11481},
  year={2022}
}

@inproceedings{lee2025timecap,
  title={Timecap: Learning to contextualize, augment, and predict time series events with large language model agents},
  author={Lee, Geon and Yu, Wenchao and Shin, Kijung and Cheng, Wei and Chen, Haifeng},
  booktitle={AAAI},
  volume={39},
  pages={18082--18090},
  year={2025}
}

@article{wang2024news,
  title={From news to forecast: Integrating event analysis in llm-based time series forecasting with reflection},
  author={Wang, Xinlei and Feng, Maike and Qiu, Jing and Gu, Jinjin and Zhao, Junhua},
  journal={Advances in Neural Information Processing Systems},
  volume={37},
  pages={58118--58153},
  year={2024}
}

@article{narasimhan2024time,
  title={Time weaver: A conditional time series generation model},
  author={Narasimhan, Sai Shankar and Agarwal, Shubhankar and Akcin, Oguzhan and Sanghavi, Sujay and Chinchali, Sandeep},
  journal={arXiv preprint arXiv:2403.02682},
  year={2024}
}

@inproceedings{xu2021rest,
  title={Rest: Relational event-driven stock trend forecasting},
  author={Xu, Wentao and Liu, Weiqing and Xu, Chang and Bian, Jiang and Yin, Jian and Liu, Tie-Yan},
  booktitle={WWW},
  pages={1--10},
  year={2021}
}

@article{lian2026contextual,
  title={Contextual Masking Distillation for Network Traffic Anomaly Detection},
  author={Lian, Xinglin and Zheng, Yu and Liu, Yan and Zhou, Fan and Peng, Chunlei and Gao, Xinbo},
  journal={IEEE Transactions on Information Forensics and Security},
  year={2026},
  publisher={IEEE}
}

@inproceedings{qiu2026dag,
  title={DAG: A Dual Correlation Network for Time Series Forecasting with Exogenous Variables},
  author={Qiu, Xiangfei and Zhu, Yuhan and Li, Zhengyu and Wu, Xingjian and Yang, Bin and Hu, Jilin},
  booktitle={ICML},
  year={2026}
}

@article{liang2024exploring,
  title={Exploring large language models for human mobility prediction under public events},
  author={Liang, Yuebing and Liu, Yichao and Wang, Xiaohan and Zhao, Zhan},
  journal={Computers, Environment and Urban Systems},
  volume={112},
  pages={102153},
  year={2024},
  publisher={Elsevier}
}

@inproceedings{ge2025t2s,
  title={T2S: high-resolution time series generation with text-to-series diffusion models},
  author={Ge, Yunfeng and Li, Jiawei and Zhao, Yiji and Wen, Haomin and Li, Zhao and Qiu, Meikang and Li, Hongyan and Jin, Ming and Pan, Shirui},
  booktitle={IJCAI},
  pages={5208--5216},
  year={2025}
}

@inproceedings{guan2025timeomni,
  title={Timeomni-1: Incentivizing complex reasoning with time series in large language models},
  author={Guan, Tong and Meng, Zijie and Li, Dianqi and Wang, Shiyu and Yang, Chao-Han Huck and Wen, Qingsong and Liu, Zuozhu and Siniscalchi, Sabato Marco and Jin, Ming and Pan, Shirui},
  booktitle={ICLR},
  year={2026}
}

@inproceedings{ge2024moment,
  title={A moment cross predictor for non-stationary mobile traffic forecasting},
  author={Ge, Yunfeng and Zhang, Yingxin and Shi, Keyi and Li, Hongyan},
  booktitle={ICCC},
  pages={2059--2064},
  year={2024},
  organization={IEEE}
}

\appendix


\section{Datasets}

\subsection{Detailed Dataset Descriptions}
\label{datasetdescription}
Seven datasets are used to evaluate \method, including one synthetic dataset and Six real-world datasets. Statistical descriptions of the multimodal event-aware datasets are presented in Table~\ref{tab:7dataset_comparison}, while the event embedding spaces for each dataset are visualize in Figure~\ref{fig:7embeddingsVisualization}.
All datasets are publicly available, and the construction methods are described below:

\noindent\textbf{Synthetic Dataset}. To assess \method's event-aware capabilities, we construct a synthetic dataset to simulate how events manifest as shifts in time series patterns. It includes four waveform types: ``sinusoidal, triangular, sawtooth, and near-square waves''.
Each waveform is perturbed with varying noise to mimic realistic signals.
Segments are annotated with a brief description, such as ``sine wave,'' ``triangle wave with noise,'' or ``sawtooth wave with spikes.''
The textual component is obtained by first prompting GPT-4o to describe each waveform segment, followed by embedding generation using OpenAI’s \texttt{text-embedding-3}.
These paired time series and textual embeddings form a comprehensive synthetic dataset.
As shown in Table~\ref{tab:7dataset_comparison}, the Synthetic Dataset comprises a balanced number of samples with an average description length of 2.14 words.
As shown in Figure~\ref{fig:7embeddingsVisualization}, the t-SNE visualization of the text embeddings reveals a clear separation in the embedding space, demonstrating the dataset's effectiveness in simulating multimodal event scenarios.

\noindent\textbf{Atmospheric Physics–Weather Events Dataset}. This dataset pairs atmospheric measurements from the Beutenberg weather station (Jena, Germany) with natural language descriptions from a local meteorological service.
These descriptions cover various aspects of the weather, including cloud conditions such as broken clouds, partly sunny skies, and passing clouds, as well as perceived comfort levels and relative humidity.
These descriptions are then embedded into vector representations using  OpenAI’s text embedding model and aligned with 10-minute resolution time series.
For the time series modality, Vapor Pressure Deficit is used as the primary signal, capturing air dryness and correlating with the textual attributes.
As shown in Table~\ref{tab:7dataset_comparison}, the Atmospheric Physics–Weather Events Dataset has the second largest number of samples and exhibits the finest sampling rate. The average event description length is 3.64 words.
Due to the presence of both discrete and continuous attributes, event types are not easily categorized.
Nevertheless, the t-SNE plot in Figure~\ref{fig:7embeddingsVisualization} shows distinct clusters, revealing latent semantic structure. This dataset offers a fine-grained, multimodal benchmark for event-aware forecasting.

\noindent\textbf{Traffic–Public Events Dataset}. Following the methodology in~\cite{aliang2024exploringa}, this dataset captures how regional traffic is influenced by public events. We collect hourly green taxi drop-off data near the Barclays Center (NYC) from January to June 2015, alongside official event schedules from the same period. Each event description is aligned with time series segments where traffic is expected to be impacted. For non-event periods, we assign a $\textsc{null}$ event to maintain modality alignment.
As shown in Table~\ref{tab:7dataset_comparison}, the Traffic–Public Events Dataset contains the fewest samples, with an average description length of 4.21 words, mainly reflecting full event names. Since events are unconstrained and belong to an open set, categorical statistics are not provided.
Nevertheless, Figure~\ref{fig:7embeddingsVisualization} shows that despite variations in textual embeddings, the t-SNE visualization reveals several well-defined clusters in the center, indicating that the latent representations of event-related segments exhibit meaningful separability.

\noindent\textbf{Temperature–Rainfall Events Dataset~\cite{alee2025timecapa}}.
The Temperature–Rainfall Dataset comprises hourly temperature time series data collected over a five-year period from several major U.S. cities, including New York, San Francisco, and Houston, accompanied by binary rainfall event labels. These labels indicate the presence or absence of rainfall at each time step.
As shown in Table~\ref{tab:7dataset_comparison}, the textual event descriptions are limited to a single token, either $\textsc{true}$ or $\textsc{false}$, reflecting the binary nature of the rainfall labels. These binary values are directly encoded as textual event embeddings using simple one-hot encoding.
Consequently, as illustrated in Figure~\ref{fig:7embeddingsVisualization}, the t-SNE visualization of the embedding space reveals two separated binary clusters, corresponding to the binary event categories. This separation is consistent illustrated across all three cities, demonstrating the coherence of the textual embeddings despite geographic variation.

\noindent\textbf{Traffic-News Events Dataset~\cite{awang2024newsa}}.
The Traffic–News Events Dataset~\cite{awang2024newsa} pairs traffic load records with textual news events. Traffic news contains that a broad range of external factors, including holidays, weather conditions, and other impactful occurrences.
As shown in Table~\ref{tab:7dataset_comparison}, this dataset features the longest textual descriptions among all datasets, with an average length of 280 words per event. The textual embeddings are generated using OpenAI’s \texttt{text-embedding-3} model.
The event embedding space is visualized in Figure~\ref{fig:7embeddingsVisualization}. Despite the richness of the content, the embeddings exhibit weakly differentiated clusters. This suggests that long news descriptions may dilute core event semantics. This lack of semantic compactness in embedding space poses challenges in event-aware modeling.

\subsection{Case Study of the Traffic-News Events Dataset}
\label{datasetanalysis}
\begin{figure*}
    \centering
    \includegraphics[width=1\linewidth]{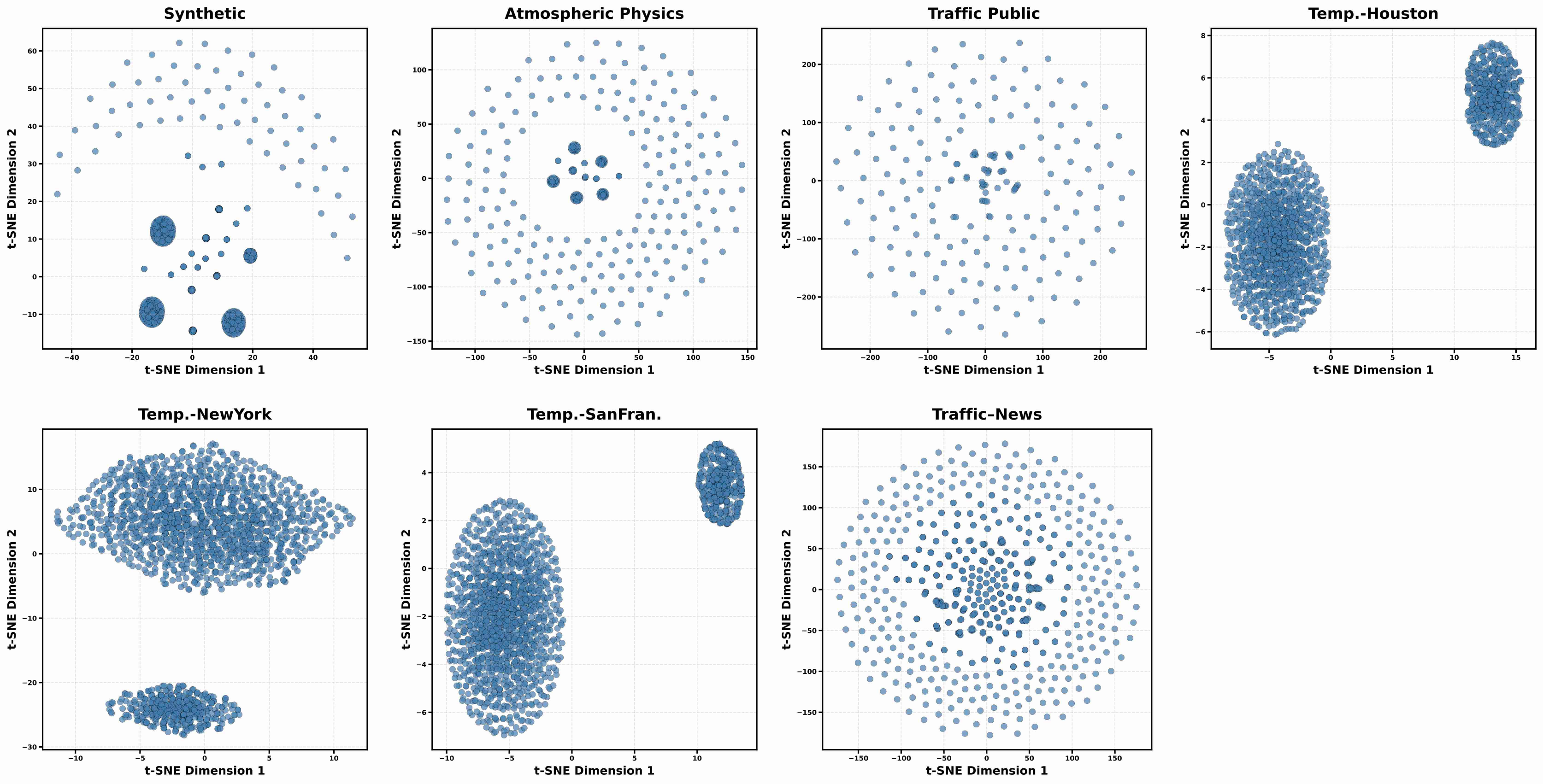}
    \caption{Comparison of event embedding space. Across most datasets, textual event embeddings form well-defined clusters, indicating semantically distinct and well-separated event representations. In contrast, the Traffic–News dataset exhibits weak clustering and substantial overlap among embeddings, suggesting limited semantic separability in its event descriptions.}
    \label{fig:7embeddingsVisualization}
\end{figure*}

The Traffic–News Events Dataset yields suboptimal performance in the main experiments, which can be primarily attributed to limitations in the quality and structure of its textual modality. To quantitatively assess the contribution of textual information to forecasting accuracy, we introduce the $\Delta_{\mathcal{V}}\textit{J-FTSD}$ metric, as detailed in Appendix~\ref{app:preditable}.
As shown in Figure~\ref{fig:newsdisplay}, approximately $87.3\%$ of the event descriptions consist of background news content that lacks direct relevance to the corresponding time series behavior. Such redundancy impairs the ability to establish clear temporal and semantic alignment between textual events and time series segments.
Furthermore, the OpenAI \texttt{text-embedding-3} model fails to produce semantically coherent clusters for these background reports, as illustrated in Figure~\ref{fig:7embeddingsVisualization}. This observation suggests a lack of structure in the resulting embedding space, which introduces additional challenges for models attempting to exploit these embeddings for downstream forecasting tasks.
Consequently, these findings highlight that the informativeness and interrelation of textual events are pivotal for event-aware forecasting.
The inclusion of textual data alone does not guarantee performance gains; rather, the utility of such data depends on its semantic clarity and its correlation with the underlying time series dynamics.
These findings motivate more deliberate dataset design and event annotation for event-aware time series modeling.
\begin{figure}[t]
    \centering
    \includegraphics[width=1\linewidth]{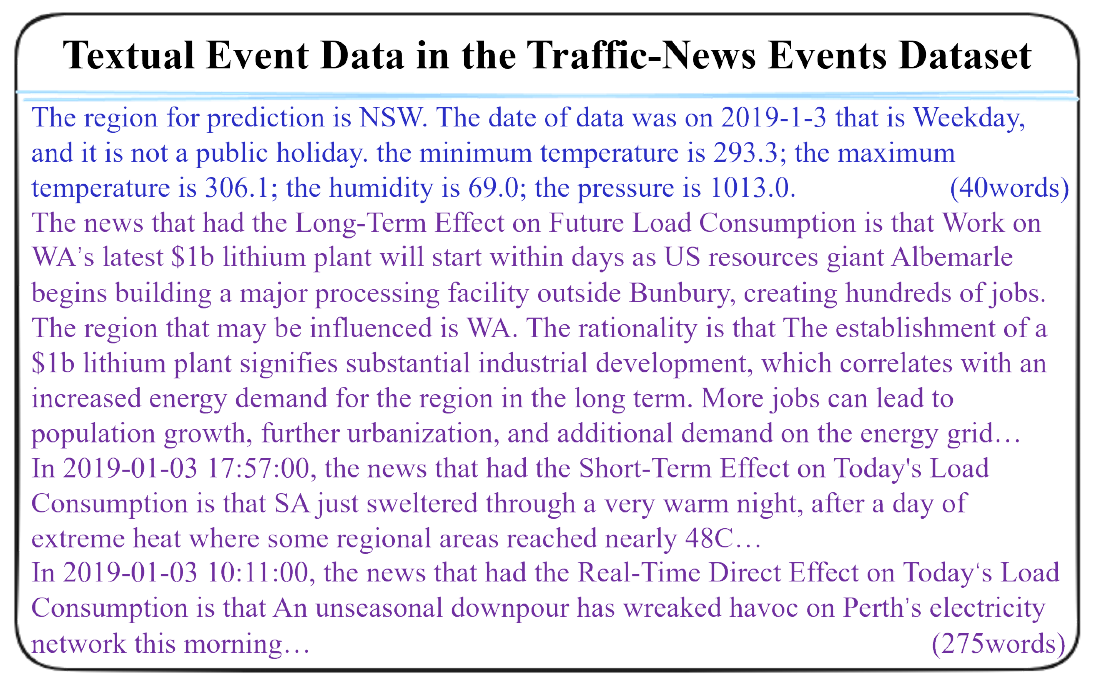}
    \caption{Example of a textual event, illustrating excessive verbosity and weak correlation with the traffic series.}
    \label{fig:newsdisplay}
\end{figure}

\begin{table*}
    \centering
    \begin{tabular}{lcccc}
        \toprule
        \textbf{Datasets} & \textbf{Synthetic} & \textbf{Atmosphere} & \textbf{Traffic–Pub.} & \textbf{Temp.Houston} \\
        \midrule
        \textbf{Timestamps} & 26,280 & 52,743 & 4,704 & 45,216 \\
        \textbf{Frequency} & 24 points per wave & 10min & hourly & hourly \\
        \textbf{Duration} & 1095 waves & 2014-01 to 2015-01 & 2015-01 to 2015-06 & 2012-10 to 2017-11 \\
        \textbf{Main Domain} & Synthetic waveform & Physics record & Traffic & Weather \\
        \textbf{Auxiliary Domain} & Waveform description & Weather activity & Public activity & Rainfall activity \\
        \textbf{Event Distribution} & \makecell{Sin(35.6\%):Tri(32.2\%)\\Sqr(24\%):Saw(8.2\%)} & --- & --- & \makecell{Rain (24.26\%)\\Not rain (75.74\%)}\\
        \textbf{Desc. Length} & 2.14 words & 3.64 words & 4.21 words & 1 word \\
        \midrule
        \textbf{Datasets} & \textbf{Temp.NewYork} & \textbf{Temp.SanFran.} & \textbf{Traffic–News} \\
        \midrule
        \textbf{Timestamps} & 45,216 & 45,216 & 35,088 \\
        \textbf{Frequency} & hourly & hourly & 30min \\
        \textbf{Duration} & 2012-10 to 2017-11 & 2012-10 to 2017-11 & 2019-01 to 2020-12 \\
        \textbf{Main Domain} & Weather & Weather & Traffic \\
        \textbf{Auxiliary Domain} & Rainfall activity & Rainfall activity & Social media activity \\
        \textbf{Event Distribution} & \makecell{Rain (24.58\%)\\Not rain (75.42\%)} & \makecell{Rain (30.94\%)\\Not rain (69.06\%)} & --- \\
        \textbf{Desc. Length} & 1 word & 1 word & 280.04 words \\
        \bottomrule
    \end{tabular}
    \caption{
    Statistical comparison of datasets. Time series attributes are reported, including the number of timestamps, data frequency, and total duration. The main domain refers to the time series domain; the auxiliary domain refers to the corresponding textual event domain; the proportion of event distributions and the average length of event descriptions are also reported.
    }
    \label{tab:7dataset_comparison}
\end{table*}

\subsection{Quantifying Predictability Gains from Textual Events}
\label{app:preditable}
To rigorously quantify textual events' contribution to forecasting performance, we propose $\Delta_{\nu} \textit{J-FTSD}$, which measures the improvement in $\textit{J-FTSD}$ when incorporating versus excluding textual event inputs. $\textit{J-FTSD}$ evaluates alignment between real and generated joint distributions of time series and associated metadata~\cite{anarasimhan2024timea}, where lower values indicate better fidelity.

Let $\mathcal{D}_r = \{(\mathbf{x}_s, \mathbf{c}_s)\}_{s=1}^N$ denote the ground truth dataset with $N$ time series and textual event pairs, and let $\mathcal{D}_g^{v}= \{(\mathbf{x}_s^v, \mathbf{c}_s)\}_{s=1}^N$ be its perturbed version under noise level $\nu \in \mathcal{V}$, which proxies uncertainty outputs of generative models. We define the event-removed versions $\mathcal{D}_r^{'} = \{(\mathbf{x}_s, \mathbf{c}_{\text{noise}})\}$ and $\mathcal{D}_g^{'v} = \{(\mathbf{x}_s^v, \mathbf{c}_{\text{noise}})\}$, where event context is replaced with noise $c_{\text{noise}} \sim \mathcal{N}(0,\mathbf{I})$. Then,
\begin{small}
\begin{equation}
\Delta_{\mathcal{V}}\textit{J-FTSD} = \frac{1}{|\mathcal{V}|} \sum_{\nu \in \mathcal{V}} \left[ \textit{J-FTSD}(\mathcal{D}_r^{'}, \mathcal{D}_g^{'v}) - \textit{J-FTSD}(\mathcal{D}_r, \mathcal{D}_g^{v}) \right]
\end{equation}
\end{small}

\begin{figure}
    \centering
    \includegraphics[width=1\linewidth]{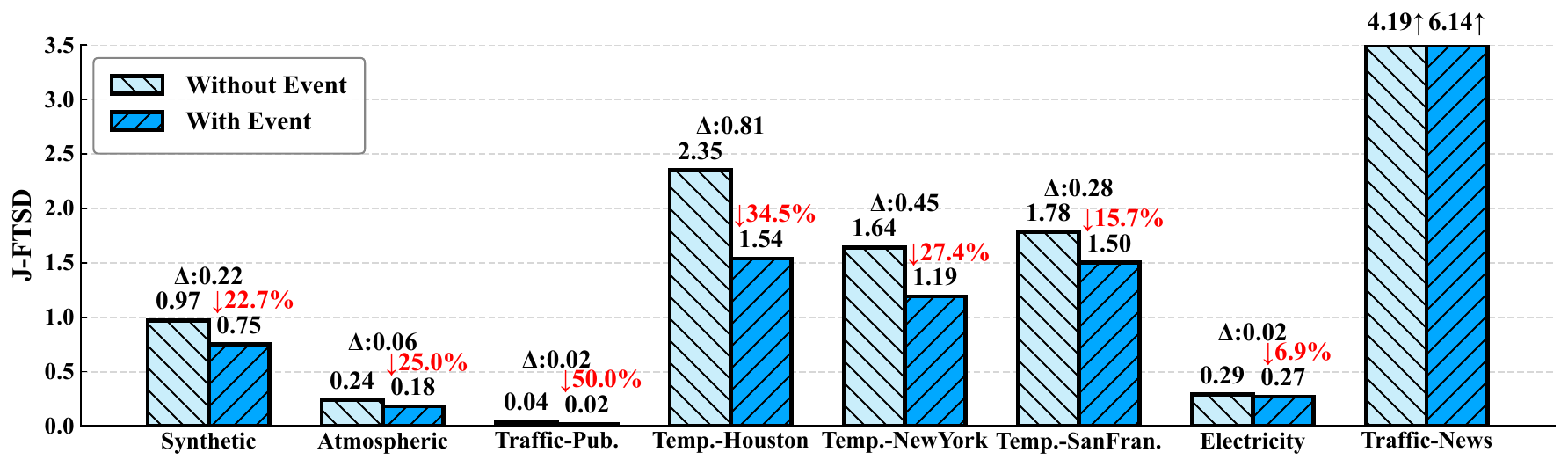}
    \caption{The $\Delta_{\mathcal{V}}\textit{J-FTSD}$ metric quantifies predictability gains for event-aware non-stationary time series datasets.}
    \label{fig:jftsd}
\end{figure}

Figure~\ref{fig:jftsd} shows that textual events improve predictive alignment across most datasets. However, improvements are negative for the Traffic–News Events dataset, consistent with the results observed in the main experiments. Dataset analysis shows that the textual descriptions are excessively redundant, hindering efficient alignment. See the Appendix~\ref{datasetanalysis} for details. Results emphasize that informative and relative textual events are crucial for improving multimodal event-aware time series forecasting performance.

\section{Detailed Design}

\subsection{Multimodal U-shaped Diffusion Transformer}
\label{app:mudit}
\begin{figure}[t]
    \centering
    \includegraphics[width=0.9\linewidth]{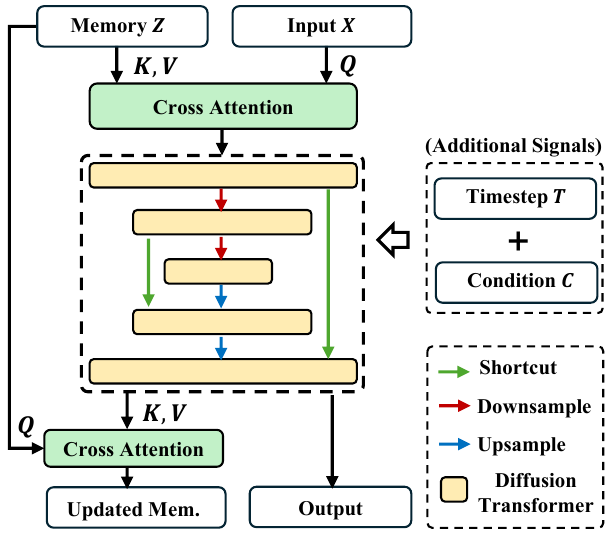}
    \caption{Multimodal U-shaped Diffusion Transformer. Aligns events with multi-resolution temporal patterns.
    }
    \label{appfig:muditstructure}
\end{figure}

Textual events and temporal dynamics often exhibit representational misalignment, especially when events have multi-resolution impacts ranging from coarse-grained trends to fine-grained surges. To address this challenge, we propose the Multimodal U-shaped Diffusion Transformer (M-U-DiT) in Figure~\ref{appfig:muditstructure}, inspired by the U-Net model~\cite{aronneberger2015ua}. M-U-DiT leverages down-sampling, up-sampling, and skip connections to bridge textual event semantics with multi-resolution temporal components.
M-U-DiT takes current noisy input $\mathbf{x}_{s}^{t}$ (or $\mathbf{n}_{s}^{t}$ during sampling), previous latent state $\mathbf{Z}_{s-1}$, flow matching timestep $t$, and textual condition $\mathbf{c}_s$, then outputs estimated velocity field and state $\hat{\mathbf{Z}}_{s}$.

\noindent \textbf{Input}. Input processing begins with cross-attention that integrates historical information $\mathbf{Z}_{s-1}^t$ at diffusion timestep $t$ and current noisy observation $\mathbf{x}_{s}^{t}$, producing enriched representation $\mathbf{x}_e^{(0)}$ as input to the stacked M-U-DiT.
Meanwhile, diffusion timestep $t$ and textual condition $\mathbf{c}_s$ are embedded and fused to form conditioning vectors $\mathbf{g}_{*} = \phi_{*}(t) + \Phi_{*}(\mathbf{c}_s)$, where $\phi_{*}$ and $\Phi_{*}$ are embedding layers. To match feature dimensions across different M-U-DiT components, global vectors following this formulation $\mathbf{g}_{e}$, $\mathbf{g}_{b}$, and $\mathbf{g}_{d}$, are used for the encoder, bottleneck layer, and decoder, respectively.

\noindent \textbf{M-U-DiT}. We denote the $l$-th encoder layer as $\epsilon_e^{(l)}$, the single bottleneck layer as $\epsilon_b$, and the $l$-th decoder layer as $\epsilon_d^{(l)}$, where $l \in \{1, \dots, M\}$. Down-sampling and up-sampling operations, denoted as $\textit{Down-sampling}$ and $\textit{Up-sampling}$, are applied in the encoder and decoder layers.

M-U-DiT takes the temporal features $\mathbf{x}_{e}^{(0)}$ and global condition $\mathbf{g}_{*}$ as input, and processes them through the down-sampling encoder, a bottleneck layer, and the up-sampling decoder.
The computation at the $l$-th down-sampling encoder layer is defined as follows:
\begin{equation}
\tilde{\mathbf{x}}_e^{(l)}=\epsilon_e(\mathbf{x}_e^{(l-1)},\mathbf{g}_e),
\end{equation}
\begin{equation}
\mathbf{x}_e^{(l)}= \begin{cases}\textit{Down-sampling }\left(\mathbf{\tilde{x}}_e^{(l)}\right), & l<M \\ \mathbf{\tilde{x}}_e^{(l)} & l=M\end{cases}.
\end{equation}
The output of the $M$-th encoding layer $\mathbf{x}_e^{(M)}$ is fed to the bottleneck layer as $\mathbf{\tilde{x}}_b$. The bottleneck computation is $\mathbf{x}_b=\epsilon_b(\mathbf{\tilde{x}}_b,\textbf{g}_b)$.
Following this, the bottleneck output $\mathbf{x}_b$ initializes the decoder as $\mathbf{x}_d^{(0)} = \mathbf{x}_b$. The decoding computation at the $l$-th up-sampling decoder layer is defined as follows:
\begin{equation}
\mathbf{\tilde{x}}_d^{(l)}= \begin{cases}\textit{Up-sampling}\left(\mathbf{x}_d^{(l-1)}\right), & l>1 \\ \mathbf{x}_d^{(l-1)} & l=1\end{cases},
\end{equation}
\begin{equation}
\mathbf{x}_d^{(l)}=\epsilon_d(\mathbf{\tilde{x}}_d^{(l)}+\mathbf{x}_e^{(l)},\mathbf{g}_d).
\end{equation}
\noindent \textbf{Output}. The final decoder layer output $\mathbf{x}_{d}^{(M)}$ predicts the velocity field. Updated latent state is derived via cross-attention between output and original latent state $\mathbf{Z}_{s-1}^t$.

\begin{figure}
    \centering
    \includegraphics[width=\linewidth]{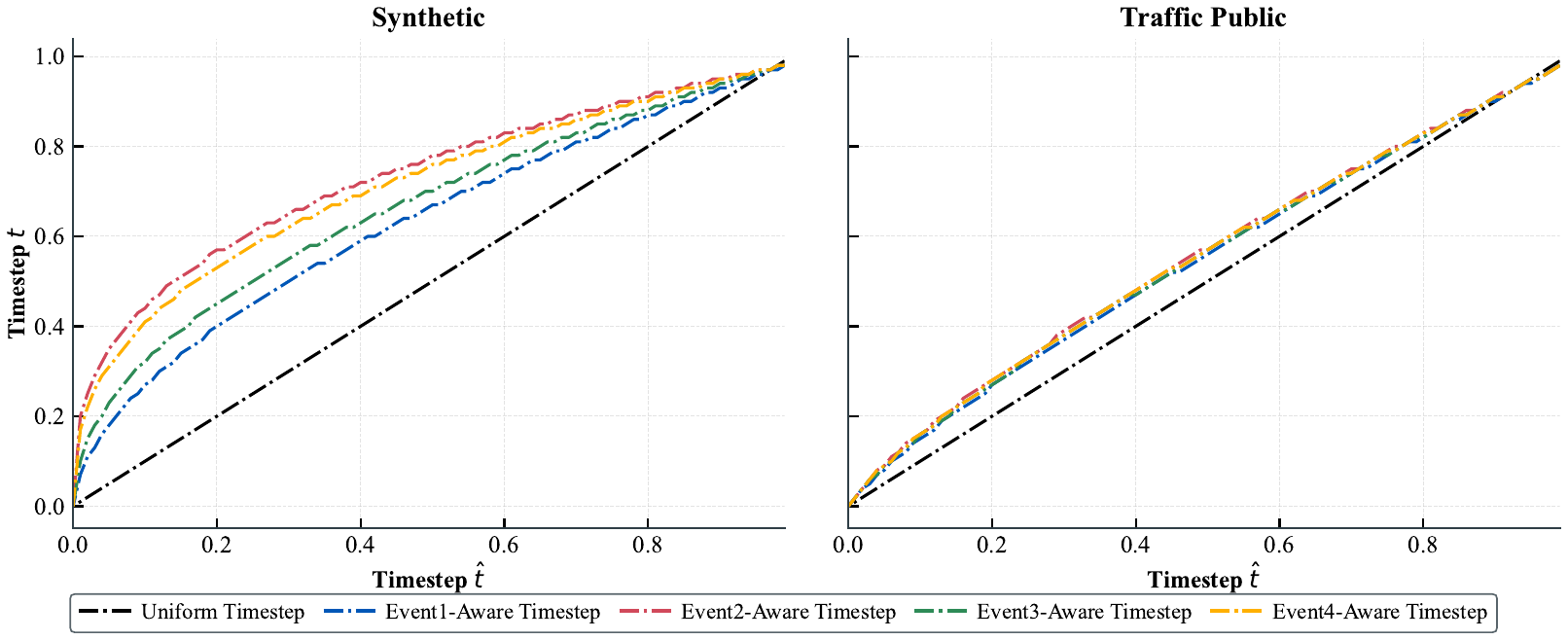}
    \caption{Learned event-aware diffusion timestep. Visualization of the learned timestep reparameterization $t=\phi(\hat{t},\mathbf{c})$ across event categories on Synthetic and Traffic-Public Datasets.}

    \label{fig:timestepcombined}
\end{figure}

\subsection{Event-Aware Timestep Validity via Learned Schedule Visualization}
\label{app:eventawaredifferenttimestep}
Figure~\ref{fig:timestepcombined} visualizes the learned event-aware reparameterization from the base uniform timestep index $\hat{t}$ to the effective diffusion timestep $t$, providing direct evidence that our scheduler is \emph{event-dependent} rather than event-agnostic.
Under the same $\hat{t}$, the mapping $\phi(\hat{t},\mathbf{c})$ redistributes sampling locations in a clearly nonlinear transformation, and the resulting curves exhibit systematic shape variations across event categories within the same dataset.
This phenomenon is consistently observed in both datasets, indicating that the scheduler is being reliably modulated by event semantics.

These learned mappings align with our motivation: event-induced shifts in signal magnitude and variance change the effective SNR and thus the denoising difficulty even at identical nominal timestep.
Consequently, a uniform timestep schedule can be suboptimal under heterogeneous event conditions.
By learning an event-conditioned reparameterization, the model allocates denoising budget toward difficulty-critical stages for each event type, leading to a more balanced training signal across diffusion stages.
Importantly, this qualitative evidence is consistent with our quantitative ablations, where replacing uniform timestep with event-aware timestep yields consistent improvements across datasets and metrics, supporting the necessity of the event-aware timestep.

\section{Training Details}
\label{app:settings}
All experiments are conducted on NVIDIA 4090 GPUs running Ubuntu 22.04 LTS with 64GB of system RAM. The software environment are based on Python 3.11.10 and PyTorch 2.5.0 with CUDA 12.1.1. Complete software and package specifications are provided in the code repository.

Models are trained for a maximum of $1000$ epochs with a batch size of $64$ with random seed $42, 43, 44$ over three times. Due to varying sampling granularities across datasets, most utilized an input length of $168$, an output length of $92$, and a segment slice size of $24$. However, for the Atmospheric dataset, the input and output lengths are set to $288$ with a segment size of $144$, while the Traffic News dataset utilized an input and output length of $192$ with a segment size of $48$.
Early stopping is applied every $5$ epochs with a patience of $5$. Input time series are z-score normalized. We use dynamic dropout, linearly decreasing the dropout rate from $0.6$ to $0.05$ over training.
We employed the AdamW optimizer with an initial learning rate $2 \times 10^{-4}$ and a weight decay of $1 \times 10^{-3}$. The optimizer is configured with $\beta_1 = 0.9$ and $\beta_2 = 0.999$, and the learning rate followed a OneCycleLR schedule. The learning rate is warmed up from $1 \times 10^{-5}$ to the peak over the first 20\% of training steps. It then linearly decayed to a final value of $1 \times 10^{-7}$ over the remaining steps.
The flow matching training procedure used $100$ sampling steps, while the sampling procedure employed $50$ sampling steps. The model architecture consists of $3$ encoder layers, $1$ bottleneck layer, and $3$ decoder layers. Both downsampling and upsampling rates are set to $2$. The historical state dimension is $48$, the hidden dimension is $256$, the textual embedding dimension is $128$, and the fully connected layer dimension is $1024$.

\section{More Related Work}
In the financial domain, events have been shown to enhance forecasting tasks by leveraging deep learning techniques to extract sentiment, descriptive, and semantic features from textual event data. These methods can be classified into four lines, including \ding{182} deep learning methods~\cite{aliu2018hierarchicala,aemami2023modalitya}, \ding{183} reinforcement learning methods~\cite{awang2021deeptradera}, \ding{184} generative methods~\cite{aduan2022factorvaea,axia2024marketa}, and \ding{185} large language model based methods~\cite{aliu2025llm4ftsa,axiao2025retrievala,ali2024marsa,azhang2025camefa,ayang2025learninga}.
Deep learning methods, such as REST~\cite{axu2021resta} utilizes a long short-term memory network to model the impact of newly emerged events on stock price fluctuations, while~\cite{aemami2023modalitya} introduces a modality-aligned transformer to encode financial time series and event semantics jointly.
Reinforcement learning methods, such as DeepTrader~\cite{awang2021deeptradera}, incorporate macroeconomic events into trading strategies through policy adaptation. Generative methods, such as FactorVAE~\cite{aduan2022factorvaea} disentangle event factors to improve forecasting.
Recently, LLM-based financial time series forecasting includes prompt adaptation~\cite{anie2024surveya,aliu2025llm4ftsa}, retrieval-augmented generation~\cite{axiao2025retrievala}, and generative modeling~\cite{ali2024marsa}. Such approaches aim to construct representations to estimate future dynamics more effectively.
Prior work assumes events are only observed historically and learns latent event-conditioned factors.
In contrast, \method targets the event-known forecasting settings, where future scheduled events are available at forecast time and may trigger abrupt demand shifts—for example, planned sports competitions that alter traffic patterns or upcoming promotional campaigns that induce sales spikes. By conditioning on these known events, \method models time series dynamics in a temporally aligned and causal manner, addressing a largely underexplored yet practically critical forecasting regime.
Existing event-aware methods usually use events as auxiliary historical context, leaving the modeling of the scheduled future events and their fine-grained temporal effects insufficiently explored. This motivates \method, which explicitly conditions on the scheduled fine-grained future events and captures their fine-grained interactions with the continuous time series dynamics in an autoregressive diffusion framework.

\nocite{aliang2024exploringa,alee2025timecapa,awang2024newsa,anarasimhan2024timea,aronneberger2015ua,aliu2018hierarchicala,axu2021resta,aemami2023modalitya,awang2021deeptradera,aduan2022factorvaea,axia2024marketa,anie2024surveya,aliu2025llm4ftsa,axiao2025retrievala,ali2024marsa,azhang2025camefa,ayang2025learninga}

\end{document}